%% file: 0_paper.tex
\documentclass[11pt,a4paper]{article}
\usepackage{times,latexsym}
\usepackage{url}
\usepackage[T1]{fontenc}
\usepackage{verbatim}
\usepackage{todonotes}
\usepackage{pifont}
\newcommand{\cmark}{\ding{51}}%
\newcommand{\xmark}{\ding{55}}%
\usepackage{booktabs}
\usepackage{stfloats}

\usepackage[acceptedWithA]{tacl2021v1}
\usepackage{xspace,mfirstuc,tabulary}
\usepackage{hyperref}

\newif\iftaclinstructions
\taclinstructionsfalse % AUTHORS: do NOT set this to true
\iftaclinstructions
\renewcommand{\confidential}{}
\renewcommand{\anonsubtext}{(No author info supplied here, for consistency with
TACL-submission anonymization requirements)}
\newcommand{\instr}
\fi

\usepackage{comment}

\iftaclpubformat % this "if" is set by the choice of options

\else

\fi

\title{A methodology to characterize bias and harmful stereotypes \\ in natural language processing in Latin America}

\author{
  Laura Alonso Alemany$^{1,2}$, 
  Luciana Benotti$^{1,2,3}$, Hernán Maina$^{1,2,3}$, Lucía González$^{1,2}$\\ \textbf{Mariela Rajngewerc$^{1,2,3}$, Lautaro Martínez$^{1,2}$, Jorge Sánchez$^{1}$, Mauro Schilman$^{1}$, Guido Ivetta$^{1}$}\\
  \textbf{Alexia Halvorsen$^2$, Amanda Mata Rojo$^2$, 
 Matías Bordone$^{1,2}$ and Beatriz Busaniche$^2$}
  \\
  $^1$ Sección de Computación, FAMAF, 
  Universidad Nacional de Córdoba \\
  $^2$ Fundación Via Libre,  Argentina \\
  $^3$ Consejo Nacional de Investigaciones Científicas y Técnicas, Argentina
}

\date{}

\begin{document}
\maketitle
\begin{abstract} 
%my proposal (a \textit{refrit} of the original one below):

Automated decision-making systems, specially those based on natural language processing, are pervasive in our lives. They are not only behind the internet search engines we use daily, but also take more critical roles: selecting candidates for a job, determining suspects of a crime, diagnosing autism and more. Such automated systems make errors, which may be harmful in many ways, be it because of the severity of the consequences (as in health issues) or because of the sheer number of people they affect. %(as in invisibilization). 
When errors made by an automated system affect a population more than other, we call the system \textit{biased}. 
%As in many other societal processes, bias in automated system often produces harms in historically marginalized populations, thus increasing inequality and worsening their living conditions. 
%Because of the naturalization of these systems, such 
%In many cases these harmful effects cannot be easily detected and taken care of. It becomes thus necessary to develop methodologies to inspect, detect, characterize and mitigate the possible harms that automated systems may systematically produce on societal groups.

Most modern natural language technologies are based on artifacts
obtained from enormous volumes of text using machine learning, namely language models and word embeddings. %These artifacts underlie virtually all natural language processing we use today: search, recommendation systems, etc. 
Since they are created applying subsymbolic machine learning, mostly artificial neural networks, they are opaque and practically uninterpretable by direct inspection, thus making it very difficult to audit them.

In this paper we present a methodology that spells out how social scientists, domain experts, and machine learning experts can collaboratively explore biases and harmful stereotypes in word embeddings and large language models. Our methodology is based on the following principles: 

%In this paper we provide guidance to address bias in systems that involve automatic treatment of language. After a brief picture of systematic errors in such systems, we describe some existing frameworks and approaches to deal with bias, and show how they prioritize aspects of the problem different from integrating expert knowledge. Then, we propose a methodology to characterize bias in natural language processing systems with the following principles:

\begin{enumerate}
    \item focus on the linguistic manifestations of discrimination on \textit{word embeddings} and \textit{language models}, not on the mathematical properties of the models
    \item reduce the technical barrier for discrimination experts%, be it social scientists, domain experts or other
    \item characterize through a qualitative exploratory process in addition to a metric-based approach
    \item address mitigation as part of the training process, not as an afterthought
    %\item inspect central artifacts encoding linguistic knowledge, namely, \textit{word embeddings} and \textit{language models}
\end{enumerate} 

\end{abstract}

%\input{proposed_narrative}

\input{1_introduction}  %(max 1000 pal)	
\input{2_concepts}

\input{3_previous-work} %(max 700)
\input{4_methodology}    %(max 500)
\input{5_user_story}
\input{6_summary}

\bibliography{tacl2021}
\bibliographystyle{acl_natbib}

\appendix
\input{case-studies}  %(max 3000)
%\input{prototype}

%\onecolumn

%\newpage
%\appendix
%\input{annex_case-studies}
%\section{Author/Affiliation Options as set forth by MIT Press}
%\label{sec:authorformatting}

\end{document}

%% file: 1_introduction.tex
%\todo[inline, color=red!20]{Palabras para tejer el texto: bias, stereotypes, structural discrimination}

\section{Introduction} \label{sec:introduction}

In this section, we first describe the need to involve expert knowledge and social scientists with experience on discrimination within the core development of natural language processing (NLP) systems, and then we describe the guiding principles of our methodology to do so. We finish this section with a plan for the paper that briefly describes the content of each section.  

\subsection{The motivation for our methodology}%\todo{Laura proposes this title: We need knowledge on discrimination at the core of the development process}
%\todo[inline]{The need to audit IA}
Machine learning models and data-driven systems are increasingly being used to support decision-making processes. Such processes may affect fundamental rights, like the right to receive an education or the right to non-discrimination. It is important that models can be assessed and audited to guarantee that such rights are not compromised. Ideally, a wider range of actors should be able to carry out those audits, especially those that are knowledgeable of the context where systems are deployed or those that would be affected. 

%LB: I think this text is repetitive
%\textcolor{teal}{Due to the increase in the documented harmful damages yielded by Artificial Intelligence (AI) systems in areas like health [cite], justice [cite], security [cite] and employment [cite], researchers and activists strongly request more inclusive AI algorithms. 

In this paper, we argue that social risk mitigation can not be approached through algorithmic calculations or adjustments, as done in a large part of previous work in the area. Thus, the aim is to open up the participation of experts both on the complexities of the social world and on communities that are being directly affected by AI systems. Participation would allow processes to become transparent, accountable, and responsive to the needs of those directly affected by them.

%\todo[inline, color=blue!20]{Incorporación de Amanda:}
%\textcolor{blue}{Las formas hegemónicas que actualmente desarrollan los sistemas de IA, nos conducen a escenarios cada vez más adversos para los grupos sociales históricamente marginados. A la luz de los crecientes daños documentados producidos por sistemas de aprendizaje automático en áreas como la salud, la justicia, la seguridad y el empleo, una ola creciente de investigadores y activistas demandan sistemas de IA de algoritmos inclusivos. Para mitigar los riesgos sociales, que no son abordables mediante ajustes o cálculos alogrítmicos, se busca abrir la participación tanto de las personas expertas en las complejidades del mundo social como de las comunidades que son directamente afectadas por estos sistemas. La participación permite que los procesos se vuelvan transparentes, responsables y receptivos de las necesidades de quienes se ven afectados directamente por ellos.}

%\todo[inline]{Hice un merge entre el párrafo que había acá y lo que escribió Amanda}
Nowadays, data-driven systems can be audited, but such audits often require technical skills that are beyond the capabilities of most of the people with knowledge on discrimination. The technical barrier has become a major hindrance to engaging experts and communities in the assessment of the behavior of automated systems. Human-rights defenders and experts in social science recognize and criticize the dominant epistemology that sustains AI mechanisms, but they do not master the technicalities that underlie these systems. This technical and digital barrier impedes them from having greater incidence in claims and transformations. That is why we are putting an effort to facilitate reducing the technical barriers to understanding, inspecting and modifying fundamental natural language technologies. In particular, we are focusing on a key component in the automatic treatment of natural language, namely, \textit{word embeddings} and \textit{large (neural) language models}.

\subsection{Language technologies convey harmful stereotypes}%\todo{Laura proposes this subsection}

\textit{Word embeddings} and \textit{large (neural) language models} (LLMs) are a key component of modern natural language processing systems. They provide a representation of words and sentences that has boosted the performance of many applications, functioning as a representation of the semantics of utterances. They seem to capture a semblance of the meaning of language from raw text, but, at the same time, they also distill stereotypes and societal biases which are subsequently relayed to the final applications. 

Several studies found that linguistic representations learned from corpora contain associations that produce harmful effects when brought into practice, like invisibilization, self-censorship or simply as deterrents~\cite{blodgett-etal-2020-language}. The effects of these associations on downstream applications have been treated as \textit{bias}, that is, as \textit{systematic errors} that affect some populations more than others, more than could be attributed to a random distribution of errors. This biased distribution of errors results in discrimination of those populations. Unsurprisingly, such discrimination often affects negatively populations that have been historically marginalized.

To detect and possibly mitigate such harmful behaviour, many techniques for measuring and mitigating the bias encoded in word embeddings and LLMs have been proposed by NLP researchers and machine learning practitioners~\cite{10.5555/3157382.3157584,doi:10.1126/science.aal4230}. In contrast social scientists have been mainly reduced to the ancillary role of providing data for labeling rather than being considered as part of the system's core \cite{Kapoor2022}. 

We believe the participation of experts on social and cultural sciences in the design of IA would lead to more equitable, solid, responsible, and trustworthy AI systems. It would enable the comprehension and representation of the historically marginalized communities’ needs, wills, and perspectives. 

\subsection{Our guiding principles and roadmap}

In this paper, we subscribe to the following guiding principles to  
detect and characterize the potentially discriminatory behavior of NLP systems. %address bias in automated systems, that is, systematic error, with a special focus on technologies for the automated treatment of human languages. 

In contrast with other approaches, we believe the part of the process that can most contribute to bias assessment are not subtle differences in metrics or technical complexities incrementally added to existing approaches. We believe what can most contribute to an effective assessment of bias in NLP is precisely the linguistic characterization of the discrimination phenomena. 

Thus our main goals are are:
%\todo{laura: improve phrasing}
\begin{enumerate}
    \item focus on the linguistic manifestations of discrimination, not on the mathematicl properties of the inferred models
    \item reduce the technical barrier
    \item characterize instead of diagnose, proposing a qualitative exploratory process instead of a metric-based approach
    \item mitigation needs to be addressed as part of the whole development process, and not reduced to a the bias assessment after the development has been completed
    \item allows to inspect central artifacts encoding linguistic knowledge, namely, \textit{word embeddings} and \textit{language models}
\end{enumerate}

In these principles, we keep in mind the specific needs of the Latin American region. In Latin America, we need domain experts to be able to carry out these analyses with autonomy, not relying on an interdisciplinary team or on training, since both are usually not available. %In particular, this tool involves defining the seed lexicons in an interative way and providing support for adapting the lexicons through an exploratory process. 

The paper is organized as follows. Next section develops on the fundamental concept of bias, how bias materializes in the automated treatment of language, and some artifacts that currently concentrate much linguistic knowledge and also encode stereotypes and bias, namely, \textit{word embeddings} and \textit{large (neural) language models (LLMs)}. Then,  
%and argues that they are the simplest representation of word meanings that are widely used and that embed the biases present in the data on which other NLP technologies are developed.
%Section~\ref{sec:policy} integrates a look at the public policy and regulatory aspects associated with different forms of discrimination and the emergence of bias in systems that use word embeddings. Although there are currently numerous initiatives on artificial intelligence regulations, in this paper we will deal exclusively with those that have to do with discrimination of disadvantaged groups and we will make a brief mention of the challenges in terms of intellectual property. 
section~\ref{sec:previous-work} presents relevant work in the area of bias diagnostics and mitigation, and shows how they fall short in integrating expert linguistic knowledge and making it the central issue in bias characterization. %\todo{LB: this section needs updating, it is missing several sections}
%\sout{, and also arguments why this technology underlies and shapes the behavior of current natural language technologies}. 
Section~\ref{sec:use_case} explains our methodology in a worked out user story illustrating how it puts the power to diagnose biases in the hands of people with the knowledge and the societal roles to have an impact on current technologies. We conclude with a summary of our approach, a description of our vision and next steps for this line of work. %Section~\ref{sec:methodology} presents the key aspects of our methodology for biases exploration that will be implemented in our prototype. 

In the Annex~\ref{sec:case-studies} we provide two sets of case studies in which two groups of users with different profiles applied this methodology to carry out an exploration of biases, and the observations on usability and requirements that we obtained. %The first group are data scientists with different expertise backgrounds but at least a 350 hours of training and education in data science (including the development and evaluation of machine learning models). The second group are social scientists with no previous training in programming or technical aspects of machine learning.

Our methodology uses the software we implemented, available at \url{https://huggingface.co/spaces/vialibre/edia}.

%% file: 2_concepts.tex
\section{Fundamental concepts} \label{sec:concepts}
%\todo[inline]{intro to word embeddings}
%In particular, we focus on the exploration of biases in word embeddings. 

\subsection{Bias in automated decision-making systems}

In automated decision making systems, the term \textit{bias} refers to errors that are distributed unevenly across categories in an uninteded way that cannot be attributed to chance, that is, they are not random but systematic. Such bias is considered problematic when errors create unfair outcomes, with negative effects in the world.

Although the buzzword "algorithmic bias" has been in the spotlight in the last years, most of the bias in machine learning comes from the examples used to train models. Bias can be found in any part of the machine learning life cycle: the framing of the problem, which determines which data will be collected, the actual collection, with more or less resources devoted to guarantee the quality of the data, the curation of the collected data, which can result in hugely different representations... when examples arrive to the point where different algorithms may use them, they have undergone such transformations that the room for bias left to differences in algorithms is, by comparison, residual.

\subsection{Bias encoded in NLP artifacts}

In NLP, bias is not only encoded in the classifier, but also in constructs that encode lexical meaning irrespective of the final application. In current applications, word embeddings and LLMs are widely used. They are a key component of applications such as text auto-completion or automatic translation, and have been shown to improve the performance of virtually any natural language processing task they have been applied to. The problem is that, even if their impact in performance is overall positive, they are systematically biased. Thus, even if they improve general performance, they may damage communities that are the discriminated by those biases. 

%In this section we present the basic concepts of how lexical meaning is represented in NLP systems through word embeddings and LLMs and then we discuss how biases are encoded in them and how they have been found to produce biased behaviors.
Word embeddings and LLMs are biased because they are obtained from large volumes of texts that have underlying societal biases and prejudices. Such biases are carried into the representation which are thus transferred to applications. But since these are complex, opaque artifacts, working at a subsymbolic level, it is very difficult for a person to inspect them and detect possible biases. This difficulty is even more acute for people without extensive skills in this kind of technologies. In spite of that opacity, readily available pre-trained embeddings and LLMs are widely used in socio-productive solutions, like rapid technological solutions to scale and speed up the response to the COVID19 pandemic~\cite{9671817}.
%\todo[inline, color=blue!20]{Incorporación de Amanda:}
%\textcolor{blue}{Es importante señalar que en la última década las técnicas utilizadas por detrás de los sistemas de procesamiento de lenguaje natural han atravesado transformaciones profundas. Fue hasta años recientes que la ciencia de datos aprendió a obtener o extraer directamente el significado de las palabras (mediante técnicas como los word embbedings) a través de modelos computacionales de forma efectiva. Antes de contar con estas tecnologías, lingüistas y especialistas en el área codificaban a mano grandes diccionarios y cargaban la información sobre el significado de las palabras a la red. En términos sociohistóricos esto tiene implicaciones profundas, pues nos sitúa en la bisagra de una nueva fase en la que la generación de bases de conocimiento léxico ya no requiere de la actividad humana cognitiva, sino que ahora se produce de manera automática a través de datos computarizados.}
%\todo[inline, color=blue!20]{Incorporación de Amanda:}

In the last decade, the techniques used behind natural language processing systems have undergone profound transformations. In recent years, data science has learned to extract the meaning of words (using techniques such as word embedding) through computational models in an effective way. 
Before these technologies were available, linguists and specialists in the field encoded large dictionaries by hand and uploaded information about the meaning of words to the net. In sociohistorical terms, this has profound implications, since it places us at the hinge of a new phase in which the generation of lexical knowledge bases no longer requires language experts, but now occurs automatically through models that learn from data.

\subsection{Word Embeddings}\label{sec:word-embeddings}

Word embeddings are widely used natural language processing artifacts that aim to represent the behavior of words, and thus approximate their meaning, fully automatically, based on their usage in large amounts of naturally occurring text. This is why it is necessary to have large volumes of text to obtain word embeddings.

Currently, the most popular word embeddings are based on neural networks. Training a word embedding consists in obtaining a neural network that minimizes the error to predict the behavior of words in context. Such networks are obtained with training examples artificially created from naturally occurring text. Examples are created by removing, or \textit{masking} one word in a naturally occurring sentence, then having the neural network try to guess it. Billions and billions of such examples can be automatically generated from the text that is publicly available in the internet. With such huge amount of examples, word embeddings have been able to model the behavior of words in quite a satisfactory manner, and have been successfully applied to improve the performance in many NLP tools.

The gist of word embeddings consists in representing word meaning as similarity between words. Words are considered similar if they often occur in similar linguistic contexts, more concretely, if they share a high proportion of contexts of co-occurrence. Contexts of co-occurrence are usually represented as the \emph{n} words that occur before (and after) the target word being represented. In some more sophisticated structures, contexts may include some measure of word order or syntactic structures. However, most improvements in current word representations have been obtained not by adding explicit syntactic information but by optimizing \emph{n} for the NLP task (from a few words to a few dozen words)~\cite{lison-kutuzov-2017-redefining}.  

\begin{figure}
    \centering
    \includegraphics[width=\linewidth]{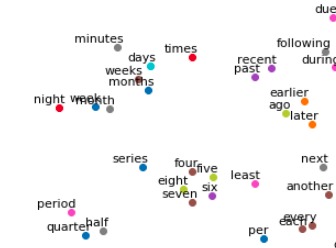}
    \caption{A representation of a word embedding in two dimensions, showing how words are closer in space according to the proportion of co-occurrences they share.}
    \label{fig:word_embedding}
\end{figure}

Once words are represented by their contexts of occurrence (in a mathematical data structure called \textit{vector}), the similarity between words can be captured and manipulated as a mathematical distance, so that words that share more contexts are closer, and words that share less contexts are farther apart, as seen in Figure~\ref{fig:word_embedding}. To measure distance, the cosine similarity is used~\cite{cinghal01}, although other distances can be used as well. The cosine similarity is useful because it is higher when words occur in the same contexts, and lower when they don't.
%\todo{here explain or cite cosine similarity}

%\todo[inline,backgroundcolor=violet!20,bordercolor=violet]{Mencionaría que es posible utilizar distintas métricas y/o que hace el cosine similarity de una manera intuitiva }

%Most Natural Language Processing systems use word representations generated by neural networks, such as word embeddings [11]. These representations are mainly used in downstream tasks such as translation, style correction, or text auto complete [18].
%It has been noted that word embeddings incorpore bias and prejudice such as associating certain professions to a gender [17]. These biases contribute to the invisibilization of minorities and stereotyping, but we are facing a shortage of methods of detection and mitigation, especially for languages apart from english [cita lit review?].

%\todo[inline]{the relevance of word embeddings and their impact in NLP}

%\todo[inline]{different word embeddings!!! important because users ask for it and it is later an important issue}

\subsection{Large Language Models (LLMs)}\label{sec:llms}

Just as word embeddings, large language models (LLMs) are pervasive in applications that involve the automated treatment of natural language. In fact, LLMs are used more than word embeddings nowadays. LLMs are artifacts that represent not individual words, but sequences of words, for example, sentences or dialogue turns. Thus, they are able to account for linguistic phenomena that escape the expressive power of word embeddings, like relations between words (i.e. agreement, co-reference), and the behavior of words in context, not in isolation. Indeed, word embeddings represent the behavior of words just by themselves, not in relation to their context.

Just like word embeddings, LLMs are obtained using neural networks and large amounts of naturally occurring text. LLMs are usually trained using the following tasks: masking a word that needs to be guessed by the neural network (similar to a fill in the blanks activity when learning a foreign language), by predicting the next words in a sentence (similar to a complete the sentence activity  when learning a foreign language), or a combination of these two. The resulting LLMs can be used for different applications, depending on their training. LLMs can be used as automatic language generators, applied to provide replies in a conversation or to complete texts after a prompt.%\todo{revise}

%\todo[inline]{I think we need here sthg about what kind of meaning (or more adequately, linguistic behaviour) llms are capturing, in contrast with word embeddings that represent the behaviour of words. We should emphasize in word embeddings that they represent the behavior of individual words}

%\todo[inline]{example of meaning that emerges because of the combination of words, as in collocations, word sense disambiguation or vagueness}

Thus LLMs capture a different perspective on the behaviour of words, a perspective that can account for finer-grained, subtler behaviors of words. For example, multi-word expressions, word senses %\todo{revise}

%% file: 3_previous-work.tex
\section{A critical survey of existing methods} \label{sec:previous-work}

%\todo[inline]{a thriving area}
In the last years the academic study of biases in language technology has been gaining growing relevance, with a variety of approaches accompanied by insightful critiques \cite{nissim-etal-2020-fair}. Most approaches build upon the experience of early proposals~\cite{10.5555/3157382.3157584,Gonen2019LipstickOA}. %\todo{could be useful to include some of the critiques to Bolukbasi that Hod includes in his original notebook} These biasses contribute to the invisibilization of minorities and stereotyping.

% Soy Mariela: perdón que puse la tabla acá pero no podría lograr ubicarla correctamente en el texto.

\begin{table*}[t]
\scriptsize
    \centering
    \begin{tabular}{l l c c c c c}
        \toprule
         \textbf{Framework} & \textbf{Reference} &\textbf{Word Embeddings }&\textbf{Language models } &\textbf{Requieres NLP} & \textbf{Mitigation} & \textbf{Counterfactuals}  \\
           &  &\textbf{Analysis}&\textbf{Analysis} &\textbf{Knowledge} & \textbf{Techniques} & \textbf{Analysis}  \\
          &  &\textbf{}&\textbf{} &\textbf{} & \textbf{Implemented} & \textbf{}  \\
         \toprule
        WordBias & \citet{Ghai2021} & \cmark & \xmark & \xmark &\xmark & \xmark  \\
        VERB & \citet{Rathore2021} & \cmark & \xmark & \cmark & \cmark & \xmark\\
        %What-if & \citet{Wexler2019} & \xmark & \xmark & \cmark & \xmark &\cmark\\
        LIT & \citet{Tenney2020} & \cmark & \cmark & \cmark & \xmark & \cmark\\

        %WEFE & & & \\
        %Fairlearn & & & \\
        %Responsibily & & & \\
        \bottomrule
    \end{tabular}
    \caption{Description of frameworks with graphical interfaces available for bias analysis of embeddings or language models. The What-if Tool is not included in the table due to is not it is not specifically targeting text data.}
    \label{tab:frameworks}
\end{table*}

In this section, we present a critical survey of previous academic work in bias assessment and we situate our proposal with respect to them. The section is organized as follows. First, we present existing tools for bias exploration. Second, we review previous work that highlights the importance of the quality of benchmarks for bias assessment. The benchmarks for bias assessment are list of words (to debias) for word embeddings and lists of stereotype and anti-stereotype sentences for language models.  
Moreover, we argue that most previous work has focused on a rather narrow set of biases and languages. Then, we discuss the risks and limitations of previous work which focuses on developing algorithms for measuring and mitigating biases automatically. Finally, we discuss the role that training data play in the process and review work that focuses on the data instead of focusing on the algorithms.

%\todo[inline,backgroundcolor=violet!20,bordercolor=violet]{Veo que quedo en comentado algo interesante, dice "Could be useful to include some of the critiques to Bolukbasi. These biasses contribute to the invisibilization of minorities and stereotyping." Lo incorporaría. También pondría alguna fortaleza del modelo si es que hay.}

%\todo[inline,backgroundcolor=pink!20,bordercolor=pink]{LB: Yo creo que este no sería el lugar para hacerlo. Habría que ver en qué subsección entra. Por el moment lo veo desconectado de la historia, no lo haría}

\subsection{Existing frameworks for bias exploration}\label{sec:existing-tools}

Multiple frameworks were developed in the last years for bias analysis. Most of them require mastery of machine learning methods and programming knowledge. Others have a graphic interface and may reach a larger audience. %that facilitates the analysis, especially for researchers without technical skills. 
%Although the possibility of accessing a graphical interface is one of the most obvious barriers to bias analysis in embeddings, other aspects such as the options present in the interfaces, the types of bias, 
%are decisive in being able to create interdisciplinary bias analysis with diverse perspectives corresponding to experts from different study areas  None of these frameworks were designed with the goal of being usable by social scientists or people without technical and programming skills in general. 
However, having a graphical interface is certainly not enough to guarantee an analysis without relying on researchers with machine learning skills. In order to create interdisciplinary bias analysis, it will be relevant that the analysis options present in the graphical interfaces, and data visualizations and manipulation are oriented toward a non-technical audience. In the following, we give a brief description of some of the frameworks with graphical interfaces available for bias analysis of embeddings or language models.

%\textcolor{violet}{
WordBias \cite{Ghai2021} is a framework that aims to analyze embeddings biases by defining lists of words. In WordBias, new variables and lists of words may be defined. This framework allows the analysis of intersectional bias. The bias evaluation is done by a score based on cosine distance between vectors and does not allow the incorporation of other metrics. Until October 2022, this framework is only available to analyze the word2vec embedding, without having the possibility to introduce other embeddings or models.%} 

%\textcolor{violet}{
The Visualizing of embedding Representations for deBiasing system (VERB) \cite{Rathore2021} is an open-source graphical interface framework that aims to study word embeddings. VERB enables users to select subsets of words and to visualize potential correlations. Also, VERB is a tool that helps users gain an understanding of the inner workings of the word embedding debiasing techniques by decomposing these techniques into interpretable steps and showing 
how words representation change using dimensionality reduction and interactive visual exploration. The target of this framework is, mainly, researchers with an NLP  background, but it also helps NLP starters as an educational tool to understand some biases mitigations techniques in word embeddings.
%VERB \cite{Rathore2021} enables bias analysis in embeddings although it is focused on bias mitigation strategies. This framework is orientated to researchers with algorithmic, linear algebra and metrics skills.%
%}

%\textcolor{violet}{
%\todo[inline]{Sacaría what-if porque no esta orientado a texto directamente, entiendo que what-if es la precuela de LIT}
What-if tool \cite{Wexler2019} is a framework that enables the bias analysis corresponding to a diverse kind of data. Although it is not focused on text data it allows this type of input. What-if tool offers multiple kinds of analysis, visualization, and evaluation of fairness through different metrics. To use this framework researchers with technical skills will be required to access the graphic interface due to is through Jupyter/ Colab Notebooks, Google Cloud, or Tensorboard, and, also, because multiple analysis options require some machine learning knowledge (e.g, selections between AUC, L1, L2 metrics). Own models can be evaluated but since it is not text-specific, it is not clear how the evaluation of words or sentences will be. This tool allows the evaluation of fairness through different metrics.%}

%\textcolor{violet}{
The Language Interpretability Tool (LIT) \cite{Tenney2020} is an open-source platform for visualization and analysis of NLP models. It was designed mainly to understand the models' predictions, to explore in which examples the model underperforms, and to investigate the consistency behavior of the models by analyzing controlled changes in data points. LIT allows users to add new datapoints on the fly,  to compare two models or data points, and provides local explanations and aggregated analysis. However, this tool requires extensive NLP understanding from the user.%}

\citet{DBLP:conf/ijcai/BadillaBP20} is an open source Python library called WEFE which is similar to WordBias in that it allows for the exploration of biases different to race and gender and in different languages. One of the focuses of WEFE is the comparison of different automatic metrics for biases measurement and mitigation. As WEFE, FairLearn \cite{Bird2020} and responsibly  \cite{responsibly} are Python libraries that enable auditing and mitigating biases in machine learning systems.  However, in order to use these libraries, python programming skills are needed as it doesn't provide a graphical interface.

%\citet{DBLP:conf/ijcai/BadillaBP20} is an open source Python library called WEFE which is similar to WordBias in that it allows for the exploration of biases different to race and gender and in different languages. One of the focuses of WEFE is the comparison of different automatic metrics for biases measurement and mitigation, however, in order to use this library python programming skills are needed as it doesn't provide a graphical interface.

%LIT provides aggregate analysis, it allows users to test local hypotheses and validate them on the dataset, to add new data points on the fly,  comparisons for two models or two data points and local explanations. However, this tool require extensive Natural Language Processing understanding from the user.

%------Fin propuesta Mariela

\subsection{On the importance of benchmarks}

\begin{figure*}[ht] 
%\begin{center}
\includegraphics[width=\linewidth]{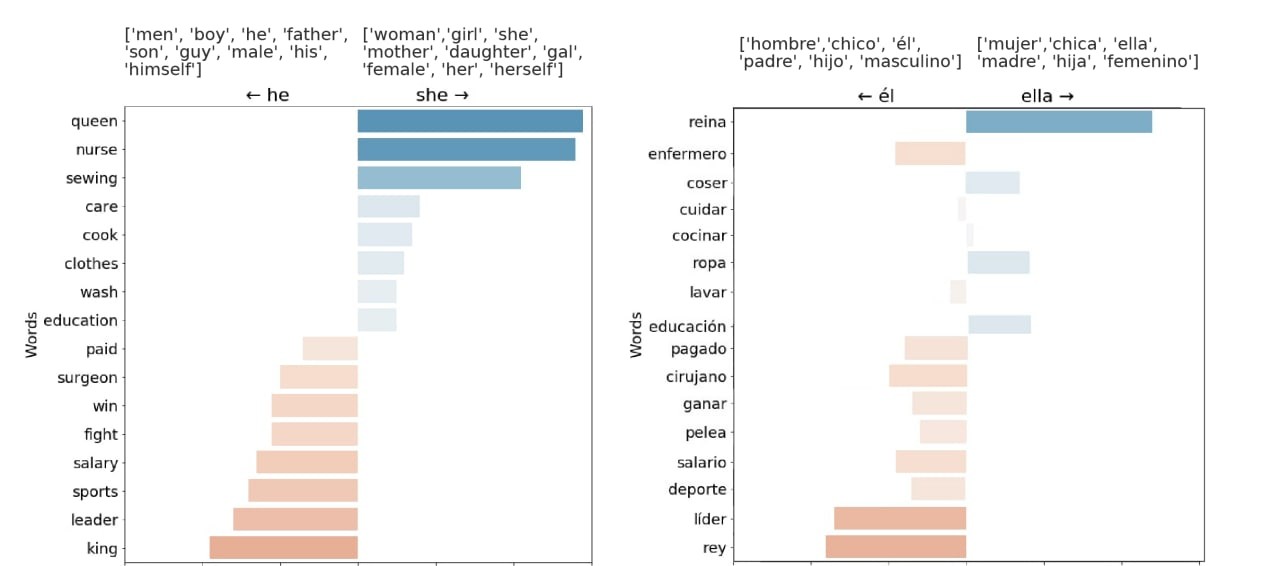}
\caption{A list of 16 words in English (left) and a translation to Spanish (right) and the similarity of their word embeddings with respect to the list of words ``\textit{woman, girl, she, mother, daughter, feminine}'' representing the concept "\textit{feminine}", the list ``\textit{man, boy, he, father, son, masculine}'' representing "\textit{masculine}", and translations for both to Spanish. The English word embedding data and training is described in~\citet{10.5555/3157382.3157584} and the Spanish in by~\cite{CaneteCFP2020}. From the 16 words of interest, in English, 8 are more associated to the concept of "\textit{feminine}", while in Spanish only 5 of them are. In particular, "\textit{nurse}" in Spanish is morphologically marked with masculine gender in the word ``\textit{enfermero}'' so, there is some degree of gender bias that needs to be taken into account to fully account for the behavior of the word.
This figure illustrates that methodologies for bias detection  developed for English are not directly applicable to other languages. Also, the figure illustrates that the observed biases depend completely on the list of words chosen.
\label{fig:english-spanish}}
%\end{center}
\end{figure*}

All approaches to assess bias on word embeddings relies on \textbf{lists of words} to define the space of bias to be explored~\cite{10.5555/3491440.3491500}. These words have a crucial impact on how and which biases are detected and mitigated, but they are not central in the efforts devoted to this task, as argued in~\citet{antoniak-mimno-2021-bad}. The methodologies for choosing the words to make these lists are varied: sometimes lists are crowd-sourced, sometimes hand-selected by researchers, and sometimes drawn from prior work in the social sciences. Most of them are developed in one specific context and then used in others without reflection on the domain or context shift. 

%LB: The next paragraph is contradicted by the one below it. 
%Nowadays, we lack systematic, transparent, and consistent documentation that details the reasons that led researchers to choose certain word lists over others, as well as a detailed discussion of word list characteristics that can cause instability.

Motivated by the aforementioned gap information and convinced that the origins and rationale underlying the lists’ selection must be explicit, tested, and documented, \citet{antoniak-mimno-2021-bad} proposed a systematic framework that enables the analysis of the sources of information and the characteristics of the instability of words in the lists, providing a detailed report about their possible vulnerabilities. %\citet{antoniak-mimno-2021-bad} provide a wide variety of methodologies to create a list of words. It is worth mentioning that these are usually developed for a specific context but then they are used in others without a serious analysis of the context change.}
They detailed three characteristics that can arise in vulnerabilities of the lists of words. In the first place, the \textit{reductionist definition}, i.e., when words are encrypted in traditional categories. This can reinforce harmful definitions or preconceptions of the categories. In the second place, the \textit{imprecise definitions}, i.e., the indiscriminate use of pre-existing datasets and over-reliance on category labels from previous works that can lead to errors. Besides, these stigmas can be enhanced with other attributes (such as gender or age) increasing the prejudice towards certain social groups. In the third place, the \textit{lexical factors}, i.e., the words' frequency and context which were selected may influence the bias analysis. 

Several recent efforts have focused on benchmark datasets consisting of pairs of contrastive sentences to evaluate bias in language models. As for methods relevant for word embeddings, these benchmarks are often accompanied by metrics that aggregate an NLP system’s behavior on these pairs into measurements of harms. \citet{blodgett-etal-2021-stereotyping}  examine four such benchmarks and apply a method—originating from the social sciences—to inventory a range of pitfalls that threaten these benchmarks’ validity as measurement models for stereotyping. They find that these benchmarks frequently lack clear articulations of what is being measured, and they highlight a range of ambiguities and unstated assumptions that affect how these benchmarks conceptualize and operationalize stereotyping. \citet{neveol-etal-2022-french} propose how to overcome some of this challenges by taking a culturally aware standpoint and a curation methodology when designing such benchmarks.

Most previous work uses word or sentence lists developed for English, or direct translations from English that do not take into account structural differences between languages~\cite{doi:10.1073/pnas.1720347115}. For example, in Spanish almost all nouns and adjectives are morphologically marked with gender, but this is not the case in English. 
Figure~\ref{fig:english-spanish} illustrates the differences in lexical biases measurements between translations of lists of words in English to Spanish over two different word embeddings in each language: the English embedding is described in ~\citet{10.5555/3157382.3157584} and the Spanish in~\citet{CaneteCFP2020}. From the 16 words analyzed, in English, 8 are more associated to the "\textit{feminine}" extreme of the bias space, while in Spanish only 5 of them are. The 3 words with different positions are ``\textit{nurse}, \textit{care} and \textit{wash}''. In particular, "\textit{nurse}" in Spanish is morphologically marked with masculine gender in the word ``\textit{enfermero}'', so it is not gender neutral. This figure illustrates two things. First, the fact that methodologies for bias detection  developed for English are not directly applicable to other languages. Second, the list of words selected to analyze bias have a strong impact on the bias that is shown by the analysis.

\subsection{On the importance of linguistic difference}\label{sec:linguistic_differences}

As illustrated by the previous example, linguistic differences have a big impact on the results obtained by the methodology to assess bias. Representing language idiosincracies is a crucial goal in characterizing biases, first, because we want to facilitate these technologies to a wider range of actors. Secondly, because to model bias in a given context or a given culture you need to do it in the language of that culture.

Different approaches have been proposed to capture specific linguistic phenomena. A paradigmatic example of linguistic variation are languages with morphologically marked gender, which can get confused with semantic gender to some extent. Most of the proposals to model gender bias in languages with morphologically marked gender add some technical construct that captures the specific phenomena. That is the case of \citet{zhou-etal-2019-examining}, who add an additional space to represent morphological gender, independent of the dimension where they model semantic gender. This added complexity supposes an added difficulty for people without technical skills.

However, it is not strictly necessary to add technical complexity to capture these linguistic complexities. A knowledgeable exploitation of word lists can also adequately model linguistic particularities. In the work presented here, we adapted the approach to bias assessment proposed by \citet{10.5555/3157382.3157584}, resorting to its implementation in the \href{https://docs.responsibly.ai/}{responsibly} toolkit. The toolkit allows for a visual and interactive exploration of binary bias as illustrated in Figure~\ref{fig:english-spanish} and explained above.
%LB: I added the previous sentence. The approach is the one explained in Figure 2. 
%\todo[inline,backgroundcolor=violet!20,bordercolor=violet]{Detallaría un poco más que hace el modelo}

\citet{10.5555/3157382.3157584}'s approach was originally developed for English, and did not envisage morphologically marked gender or the different usage of pronouns in other languages. In order to apply it to Spanish, we propose that the following considerations are necessary.

First, the extremes of bias cannot be defined by pronouns alone, because the pronouns do not occur as frequently or in the same functions in Spanish as in English. Therefore, the lists of words defining the extremes of the bias space need to be designed for the particularities of Spanish, not translated as is.

Second, with respect to the lists of words of interest to be placed in the bias space, \citet{10.5555/3157382.3157584}'s approach is strongly based on gender neutral words. However,  in Spanish most nouns and adjectives are morphologically marked for their semantic gender (as in "\textit{enfermera}", "female nurse", vs. "\textit{enfermero}", "male nurse"). To address this difference, we constructed gender neutral words resorting to patterns like: verbs, adverbs, abstract nouns, collective nouns, and adjective suffixes that are gender neutral.

\begin{figure}
    \centering
    \includegraphics[width=\linewidth]{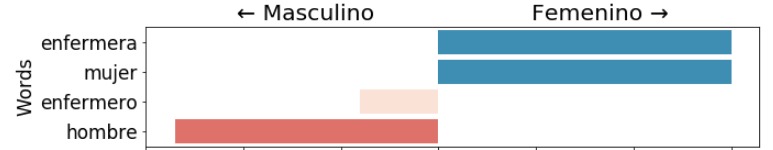}
    \caption{An assessment of how bias affects gendered words in Spanish. It can be seen that "female nurse", "\textit{enfermera}" is much more strongly associated to the feminine extreme of bias than "male nurse", "\textit{enfermero}" is associated to the masculine extreme. In contrast "woman", "\emph{mujer} is symetrically associated to female than "man", "\emph{hombre} to masculine.}
    \label{fig:gendered}
\end{figure}

Third, a proper assessment of bias for Spanish cannot be made with gender-neutral words only, because most nouns and adjectives morphologically marked for their semantic gender, or are morphologically gendered even if they have no semantics for gender (as in "\textit{mesa}", "table", which is morphologically feminine but a table is not associated to a gender). To assess bias also in that wide range of words, we constructed word lists containing both feminine and masculine versions of the same word, and compared how far they were positioned with respect to the corresponding extreme of bias. In Figure~\ref{fig:gendered} it can be seen that "female nurse", "\textit{enfermera}" is much more strongly associated to the feminine extreme of bias than "male nurse", "\textit{enfermero}" is associated to the masculine extreme. 

Summing up, putting the focus in a careful, language-aware construction of word lists has the side effect of putting in the spotlight not only linguistic differences, but also other cultural factors like stereotypes, cultural prejudices, or the interaction between different factors. Thus, in the construction of word lists, different factors need to be taken into account, not only the primary object of research that is the bias.

Researchers with experience in social domains can help understand the social and cultural complexities that underlie benchmarks constructions as argued in~\citet{blodgett-etal-2021-stereotyping}. They can also help to understand the implications of using historical datasets to train models that will be used to predict data that is markedly different from the training data~\cite{antoniak-mimno-2021-bad}.

\subsubsection{Assessing bias in word embeddings}

%LB: repeated
%Fortunately, given the importance that word embeddings have in language technologies, and the impact that biases may have, in the last years we have seen the emergence of a wide range of tools and techniques to assess bias in word embeddings and language models. 

Our core methodology to assess biases in word embeddings consists of three main parts (illustrated in Figure~\ref{fig:english-spanish} described in the previous section). The three parts are the following. 

\begin{enumerate}
    \item Defining a \textbf{bias space}, usually binary, delimited by pairs of two opposed extremes, as in \textit{male -- female}, \textit{young -- old} or \textit{high -- low}. Each of the extremes of the bias space is characterized by a list of words, shown at the top of the diagrams in Figure~\ref{fig:english-spanish}.
    \item Assessing the behaviour of \textbf{words of interest in this bias space}, finding how close they are to each of the extremes of the bias space. This assessment shows whether a given word is more strongly associated to any of the two extremes of bias, and how strong that association is. In Figure~\ref{fig:english-spanish} it can be seen that the word "\textit{nurse}" is more strongly associated to the "\textit{female}" extreme of the bias space, while the word "\textit{leader}" is more strongly associated with the "\textit{male}" extreme. 
    \item Acting on the conclusions of the assessment. The actions to be taken vary enormously across approaches, as will be seen in the next section, but all of them are targeted to \textbf{mitigate} the strength of the detected bias in the word embedding.
\end{enumerate}

This form of bias assessment characterizes words in isolation. However, the meaning of utterances depends on context, for instance in the co-occurrence of words. LLMs represent contextual meaning and we turn to them below. %\todo{develop}

\subsubsection{Assessing bias in language models}

As explained in Section \ref{sec:concepts}, LLMs represent contextual meaning. This meaning cannot be analyzed in the analytical fashion that we have seen for word embeddings. However, LLMs can be queried in terms of preferences, that is, how probable it is that an LLM will produce a given sentence. Thus, we can assess the tendency of a given LLM to produce racist, sexist language or, in fact, language that invisibilizes or reinforces any given stereotype, as long as it can be represented in contrasting sequences of words.

Methodologies to explore bias in LLMs is proposed by \citet{zhao-etal-2019-gender,sedoc-ungar-2019-role,neveol-etal-2022-french}.%\todo{include also french crows pairs, check all relevant refs are included}. %Contextual embeddings exhibit similar biases to those in word embeddings when trained on the same data~\cite{zhou-etal-2022-sense}. exploration of contextual embeddings
It consists on manually producing contrasting pairs of utterances that represent two versions of a scene, one that reinforces a stereotype and the other contrasting with the stereotype (what they call \textit{antistereotype}). Then, the LLM is queried to assess whether it has more preference for one or the other, and how much. This allows to assess how probable it is that the LLM will produce language reinforcing the stereotype, that is, how biased it is for or against the encoded stereotype.

Thus, the inspection of LLMs complements that of word embeddings.

\subsection{Race and gender, the most studied biases}

Most of the published work on biases exploration and mitigation has been produced by computer scientists based on the northern hemisphere, in big labs which have access to large amounts of founding, computing power and data. Unsurprisingly, most of the work has been carried out the English language and for gender and race biases~\cite{doi:10.1073/pnas.1720347115, blodgett-etal-2020-language, field-etal-2021-survey}. Meanwhile there are other biases that deeply affect the global south such as nationality, power and social status. Also aligned with the rest of the NLP area, work has been focused on the technical nuances instead of the more impacting qualitative aspects, like who develops the word list used for bias measurement and evaluation techniques~\cite{antoniak-mimno-2021-bad}. Since gender-related biases are one of the most studied ones, previous work has shown that the different bias metrics that exist for contextualized and context independent word embeddings only correlate with each other for benchmarks built to evaluate gender- related biases in English~\cite{DBLP:conf/ijcai/BadillaBP20}. 

%LB: Lo agregué a la subsección anterior
%\todo[inline,backgroundcolor=violet!20,bordercolor=violet]{¿Incorporar un breve comentario de \textit{morphological marking} y \textit{semantic gender}?}
English is a language where morphological marking of grammatical gender is residual, observable in the form of very few words, mostly personal pronouns and some lexicalizations like “actress - actor”. Some of the assumptions underlying this approach seemed inadequate to model languages where a big number of words have a morphological mark of grammatical gender, like Spanish or German, where most nouns or adjectives are required to express a morphological mark for different grammatical genders. The proposal from the computer science community working on bias was a more complex geometric approach, with a dimension modelling semantic gender and another dimension modelling morphological gender \cite{zhou-etal-2019-examining}. Such approach is more difficult to understand for people without a computer science background, which is usually the case for social domain experts that could provide insight on the underlying causes of the observable phenomena.
%\todo[inline,backgroundcolor=violet!20,bordercolor=violet]{Extender o dar una noción de cuál es la propuesta de les computadores, a qué se hace referencia con \textit{geometric approach, with a dimension modelling} y por qué es más complejo}

%\todo[inline,backgroundcolor=green!20,bordercolor=violet]{Laura Alonso Alemany (LAA): somewhere in this section mention the Sapir-Whorf hypothesis that language shapes concepts or concepts shape language, in relation to the possible objection that languages with morphologically marked gender are less egalitarian than languages without morphologically marked gender https://en.wikipedia.org/wiki/Linguistic\_relativity}

In this work we have explored the effects of putting the complexity of the task in constructs that are intuitive for domain experts to pour their knowledge, formulate their hypotheses and understand the empirical data. Conversely, we try to keep the technical complexity of the methodology to a bare minimum. Thus, experts from other research areas can explore and analyze bias from embeddings and language models. Learning from social science experts and including their knowledge in the analysis can help to break down dominant perspectives that reproduce inequality and marginalization, as well as work through social and structural change by developing new skills and knowledge bases. In that line, working with social science experts leads to methodological innovation in the development of computational systems.

%\todo[inline, color=blue!20]{Incorporación de Amanda:}
%\textcolor{blue}{Aprender e incluir los conocimientos de las personas con conocimiento de las fuerzas sociales y culturales ayuda a resquebrajar las perspectivas dominantes dentro de la producción de sistemas informáticos y tecnológicos que reproducen desigualdad y marginación, abriendo paso al cambio social y estructural a través del cultivo de nuevas habilidades y redes de conocimiento. La participación en este sentido resulta un mecanismo para la innovación metodológica dentro de la producción de sistemas informáticos.}

\citet{lauscher-glavas-2019-consistently} make a comparison on biases across different languages, embedding techniques, and texts. \citet{zhou-etal-2019-examining} and \citet{gonen-etal-2019-grammatical} develop 2 different detection and mitigation techniques for languages with grammatical gender that are applied as a post processing technique. %This approach requires a classifier to be trained to identify 2 gender directions: grammatical and morphological. 
These approaches add many technical barriers that require extensive machine learning knowledge from the person that applies these techniques. Therefore they fail to engage interactively with relevant expertise outside the field of computer science, and with domain experts from particular NLP applications. 
%\todo[inline,backgroundcolor=violet!20,bordercolor=violet]{Extendería con ejemplo o mini explicación cuáles son las barreras técnicas}

%what do others do about this? a focus on English, with a disregard for other languages, including language typologies. what do others do about this? Much of the posterior work assumed that accounting for such complexity required more complex approaches, which in fact hindered the involvement of experts without technical knowledge\\ Some other work \cite{gonen-etal-2019-grammatical} took more insightful approaches \todo{actually revise what they do Lucía, Laura}

\subsection{Automatically measuring and mitigating}

%\todo[inline]{a note on mitigation and bias toward technical detail}

There is a consensus \cite{field-etal-2021-survey} that what we call bias are the observable (if subtly) phenomena from underlying causes deeply rooted in social, cultural, economic dynamics. Such complexity falls well beyond the social science capabilities of most of the computer scientists currently working on bias in artificial intelligence artifacts. Most of the effort of ongoing research and innovation with respect to biases is concerned with technical issues. In truth, these technical lines of work are aimed to develop and consolidate tools and techniques more adequate to deal with the complex questions than to build a solid, reliable basis for them. However, such developments have typically resulted in more and more technical complexity, which hinders the engagement of domain experts. Such experts could provide precisely the understanding of the underlying causes that computer scientists lack, and which could help in a more adequate model of the relevant issues. We give concrete examples in Appendix~\ref{sec:case-studies}.
%\todo[inline,backgroundcolor=violet!20,bordercolor=violet]{Extendería acá: tal vez ejemplos de trabajos que comparen listas armadas por personas de la linea compu vs personas de la linea socio, o trabajos que muestren fallas en las listas.}

\citet{antoniak-mimno-2021-bad} argues that the most important variable when exploring biases in word embeddings are not the automatizable parts of the problem but the manual part, that is the word lists used for modelling the type of bias to be explored and the list of words that should be neutral. They conclude that \textit{word lists are probably unavoidable, but that no technical tool can absolve researchers from the duty to choose seeds carefully and intentionally.}

There are many approaches to "the bias problem" that aim to automatize every step from bias diagnosis to mitigation. Some of these approaches argue that when subjective, difficult decisions on how to model certain biases are involved, automating the process via an algorithmic approach is the solution~\cite{10.1145/3461702.3462536, guo-etal-2022-auto, an-etal-2022-learning, kaneko-bollegala-2021-dictionary}. \citet{DBLP:conf/acl/JiaMZC20} show that bias reduction in metrics does not correlate with bias reduction in downstream tasks. 

On the contrary, the methodology we propose in this paper hides the technical complexity. We develop on the insights of~\citet{antoniak-mimno-2021-bad} by facilitating access to these technologies to domain experts with no technical expertise, so that they can provide well-founded word lists, by pouring their  knowledge into those lists. We argue that evaluation should be carried out by people aware of  the impact that bias might have on downstream applications. Our methodology focuses on delivering a technique that can be used by such people to evaluate the bias present in text data as we explain in the next section.  

% a focus on gender and race, which most of the posterior work took as the standard task to which comparisons needed to be made, and thus disregard other biases \todo{maybe merge with previous item?}
%         \cite{du-etal-2021-assessing}

\subsection{A closer look at the training data}

%Even if mitigation directly on word embeddings has been questioned, it is still the main concern, when a more transparent option exists: to modify the texts used to train the word embeddings. 

\citet{pmlr-v97-brunet19a} trace the origin of word embedding bias back to the training data, and show that perturbing the training corpus would affect the resulting embedding bias. Unfortunately, as argued in~\citet{10.1145/3442188.3445922}, most pre-trained word embeddings that are widely used in NLP products do not describe those texts. Interestingly, \citet{dinan-etal-2020-queens} show that training data can be selected so that biases caused by unbalanced data are mitigated. Also, \citet{kaneko-bollegala-2021-dictionary} show that better curated data provides less biased models. 
%\todo[inline,backgroundcolor=violet!20,bordercolor=violet]{explicaría mejor "show that training data can be selected so that biases caused by unbalanced data are mitigated". }
%        However, it is a naive approach to assume dictionary definitions of words do not contain bias.  
%        \citet{gonen-etal-2019-grammatical} argue that this kind of techniques based on modifying the vector space do not eliminate bias but only hide it. \\

\citet{pmlr-v97-brunet19a} show that debiasing techniques have a are more effective when applied to the texts wherefrom embeddings are induced, rather than applying them directly in the already induced  word embeddings. \citet{prost-etal-2019-debiasing} show that overly simplistic mitigation strategies actually worsen fairness metrics in downstream tasks. More insightful mitigation strategies are required to actually debias the whole embedding and not only those words used to diagnose bias. However, debiasing input texts works best. Curating texts can be done automatically~\cite{gonen-etal-2019-grammatical} but this has yet to prove that it does not make matters worse. It is better that domain experts devise curation strategies for each particular case.
Our proposal is to offer a way in which word embeddings created on different corpora can be compared.

%% file: 4_methodology.tex
\section{Our methodology and prototype} \label{sec:methodology}
%{Discussion: a methodology addressing relevant aspects of bias assessment in word embeddings and large language models}%\todo{yo acá diría más bien "Work plan" o "following steps" porque siento que el research ya lo hemos hecho, ahora viene el work} 
\label{sec:discussion}
%\todo[inline]{the following paragraph originally in motivation, but I think it is better suited here}

%----------------------------
% la figura aparace aca en el codigo porque no podia ubicarla correctamente con los comandos
\begin{figure*}[ht] 
    \begin{center}
    \includegraphics[width=\linewidth]{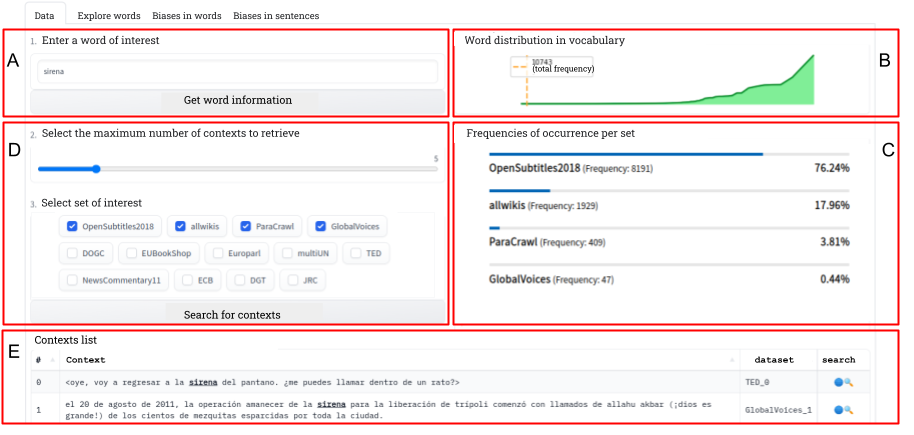}
    \caption{The EDIA visual interface, with the Data tab visible. (A) Displays the input panel for the word of interest. (B) Displays the frequency plot corresponding to all of the words in the vocabulary. The frequency of the word selected in (A) is shown in orange. (C) The panel shows the percentage of times the word selected in (A) appears in the dataset. (D) Displays the input panel for the maximum amount of contexts and dataset collections from which the user would like to get the contexts. (E) This panel shows a random sample of contexts associated with the word of interest according to the options selected in (D).
    \label{fig:visual-interface-data-tab}}
\end{center}
\end{figure*}
%------------------------------------------

In this section we summarize the main points of our methodology for bias assessment in NLP artifacts, as a concise checklist of methodological functionalities implemented in a working prototype. We also describe the EDIA prototype in detail. 

\subsection{Revisiting our guiding principles}
\label{sec:principles}

\paragraph{Focus on expertise on discrimination}, substituting highly technical concepts by more intuitive concepts whenever possible and making technical complexities transparent in the process of exploration. More concretely, by hiding concepts like "\textit{vector}", "\textit{cosine}", etc. whenever possible, for example, substituting them for the more intuitive "\textit{word}", "\textit{contexts of occurrence}", "\textit{similar}".

\paragraph{Qualitative characterization of bias}, instead of metric-based diagnosis or mitigation.%\todo{why}

\paragraph{Integrate information about diverse aspects} of linguistic constructs and their contexts.
\begin{itemize}
    \item provide context: which corpora, concrete contexts of occurrence (concordances), to get a more accurate idea of actual uses or meanings, even those that may have not been taken into account.
    \item provide information on statistical properties of words (mostly number of occurrences in the corpus, and relative frequency in different subcorpora), that may account for unsuspected behavior, like infrequent words being strongly associated to other words merely by chance occurrences.
    \item position with respect to other words in the embedding space, and most similar words.%\todo{develop} 
\end{itemize}

\paragraph{More complex bias} two-way instead of one-way, intersectionality %\todo{develop}

\paragraph{More complex representation of linguistic phenomena} word-based approaches are oversimplistic, and cannot deal with polysemy (the ambiguity or vagueness of words with respect to the meanings they may convey) or multiword expressions. That is why we need more context. Inspecting LLMs instead of word embeddings allows to account for those aspects of words. This has the added advantage of being able to inspect llms. %\todo{develop}

\subsection{The EDIA prototype for bias exploration}

 This section provides a detailed explanation of the EDIA prototype. This prototype incorporates the guiding principles described in the section~\ref{sec:principles}. EDIA is a visual interface framework for the analysis of bias in word embeddings and in LLMs that is currently available at \url{https://huggingface.co/spaces/vialibre/edia}. This prototype is hosted in huggingface so that it is easier to import pre-trained models and to offer our tool to the NLP community of practitioners around huggingface.
The EDIA visual interface is formed by four tabs.
These are: Data, Explore Words, Biases in Words , and Biases in Sentences. This framework is designed to allow the users to switch between tabs while performing the bias analysis, allowing them to utilize the exploration in one tab to feedback their analysis in the other tabs.

In the following, we will describe each tab in detail.

\begin{figure*}[t] 
    \begin{center}
    \includegraphics[width=\linewidth]{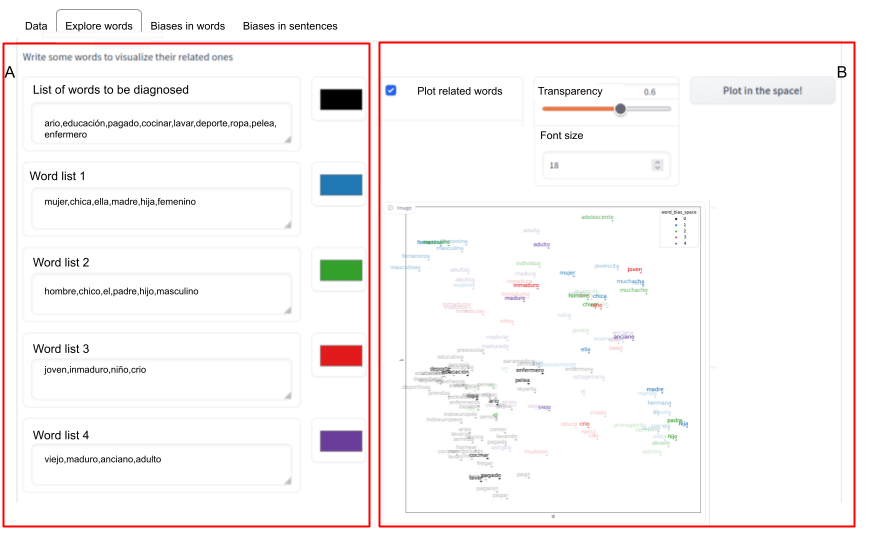}
    \caption{The EDIA visual interface, with the Explore Words tab visible. (A) Displays the input panel for the list of words of interest, and for the lists that represent each stereotype. The color next to each list (that represents a stereotype) is the one with which the words are displayed. (B) Displays the visualization panel. Three configuration options are available: word-related visualization, transparency, and font size. The list of words selected in the input panel (A) are plotted in the plane, the closeness of words indicates that they are used similarly in the corpus on which the word embedding was trained. If the "visualize related words" option is selected, the related words will appear in the same color but more transparently than the word to which they are related. 
    \label{fig:visual-interface-word-exploration-tab}}
\end{center}
\end{figure*}

\subsubsection{Data tab}
%\todo{Incluir palabra polisemia}

%\textcolor{violet}{
Huge datasets are used to train word embeddings and natural LMs. So, it is relevant to explore the way the words are used in the datasets due to biases or relations between words that appear on the corpus may be learned by the embeddings or language models. %[posibilidad de amplificar sesgos]%}

%\textcolor{violet}{
The main objectives of the Data tab are to explore the frequency of appearance of a word in the corpus as well as to explore the context of those occurrences. First, the word of interest is entered (Figure~\ref{fig:visual-interface-data-tab}A). 
Then, a graphic is visualized (Figure~\ref{fig:visual-interface-data-tab}B). This graphic depicts the frequency of appearance of each word in the corpus, while the orange line shows the frequency of the selected word in the corpus. Words with a higher frequency in the corpus (i.e., easier to find in the corpus) will have the orange line further to the right, whereas those words that hardly ever appear in the corpus will have the orange line further to the left.%}

%\textcolor{violet}{
In the default version of our tool, the objective is to explore the Word2Vec from Spanish Billion Word Corpus embedding \cite{cardellinoSBWCE}.
%\todo{Chequear}
This embedding was trained over the Spanish Unannotated Corpora \cite{canete2019} formed by 15 text collections. Figure~\ref{fig:visual-interface-data-tab}C shows the frequency of appearance of the word of interest in each of the 15 collections. The percentages on the left of each collection are calculated concerning the total frequency shown in Figure~\ref{fig:visual-interface-data-tab}B.
To get a better understanding of the word of interest in the corpus, the analysis of contexts (i.e., the sentences that appear in the corpus that contain the word of interest) is usually required. In this tab, it is possible to retrieve some of the contexts corresponding to the word of interest. To do so the maximum amount of context must be selected using the slider (five by default)  and the dataset collections from which the user would like to get the contexts must be selected (Figure~\ref{fig:visual-interface-data-tab}D). Once the "search for context" button is pressed a table showing the list of contexts appears (Figure~\ref{fig:visual-interface-data-tab}E). Each time the button is pressed a random sample of contexts are shown.%}

\subsubsection{Explore Words tab}

Both word embeddings and language models, after being trained, construct a meaning for each word in the vocabulary. This tab aims to analyze the meaning constructed by the model for a set of selected words of interest.%} 
%\textcolor{violet}{
This tab enables the visualization of a list of words of interest and lists of words that represent different stereotypes in a 2-dimensional space. 
%Each group of words is associated with a color. Furthermore, words that the model associates with the selected words of interest may also be included in the visualization.

The Explore Words tab is formed by two main panels (~\ref{fig:visual-interface-word-exploration-tab}A and B). The input panel on the left (Figure~\ref{fig:visual-interface-word-exploration-tab}A) is where the user enters the different lists of words. The rectangle next to each group of words indicates the color with which each list of words is going to be visualized. On the right (Figure~\ref{fig:visual-interface-word-exploration-tab}B), the graphics are visualized. In this panel, three configuration options are available: word-related visualization, transparency, and font size. Once the selections are made and the "Visualize" button is pressed, the visualization of the group of words appears. If the "visualize related words" option is selected, the related words appear in the same color but more transparently than the word to which they are related. The closeness of words in the 2-dimensional chart indicates that these are used similarly in the corpus on which the word embedding was trained. %}
%\todo[inline]{No estoy segura de que este bien el párrafo que sigue a continuación. Si es correcto y decidimos incluirlo podría cambiar el tipo de sesgo porque este no se que pondría en las listas de palabras, quería mostrar que se podían ver cercanías a conceptos que no necesariamente tengan dos opciones}
%\textcolor{violet}{
%With the Words Exploration tab a first approach to word embedding biases may be explored. For example, if we want to analyze the presence of gender bias in the word embedding according to professions we can complete the “list of words of interest” with a list of professions (e.g., ), the “word list 1” with words associates to female (e.g.,), the “word list 2” with words associates to non-binary  (e.g.,) and the “word list 3” with words associates to male (e.g.,). Then, we can visually explore if each selected profession is closer to any of the words-list gender representations.%}

%\todo[inline]{Lo que sigue (Words’ Bias tab) le falta, no esta como para que lo lean}

\begin{figure*}[ht] 
    \begin{center}
    \includegraphics[width=\linewidth]{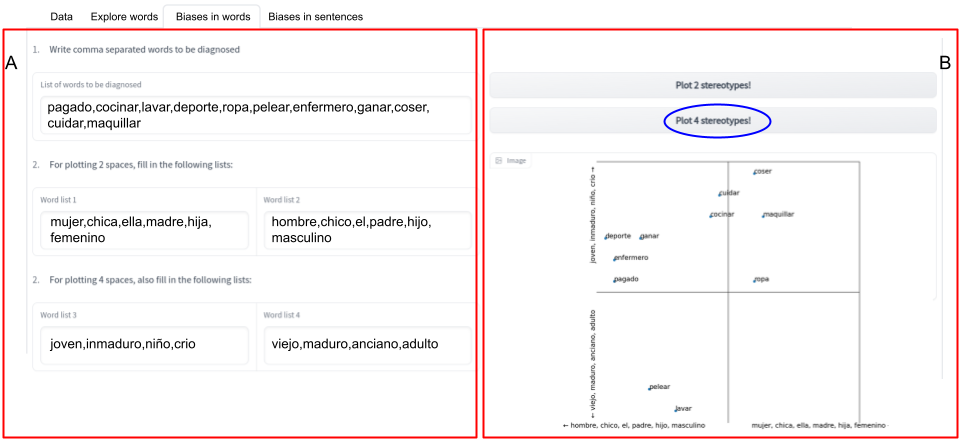}
    \caption{The EDIA visual interface, with the "Biases in Words" tab visible. (A) Displays the input panel. In t (B) Displays the visualization panel. In this figure, we show the graphic that appears after selecting the "Plot 4 Stereotypes!" option. The words of interest are plotted on a 2-dimensional graphic. Here, the x-axis represents the Word Lists 1 and 2 stereotypes. Words of interest that are plotted more to the right are more related to Word List 1 stereotypes, and words of interest that are plotted more to the left are more related to Word List 2. Furthermore, the y-axis in this graphic represents Word Lists 3 and 4. Words that are plotted more above mean that they are more related to Word List 3, whereas words that are plotted more below mean that they are more related to Word List 4.
    \label{fig:visual-words-bias}}
\end{center}
\end{figure*}

\subsubsection{Biases in Words tab}

The main objective of the Biases in Words tab is to analyze bias in a set of words of interest from a word embedding. Each type of bias to be analyzed is considered as having two possible extremes (e.g., for age the two extremes can be old and young). Each extreme is represented by a list of words.

The Biases in Words tab enables the study of up to two types of bias, allowing an intersectional analysis. This tab is formed by two panels: the input and the visualization panels. The input panel (Figure~\ref{fig:visual-words-bias}A) enables the entry of the list of the words of interest as well as the list of words that represent each extreme of bias (or stereotype) to be analyzed. The visualization panel (Figure~\ref{fig:visual-words-bias}B) is where the graphics are displayed. Two graphics are available: one that enables the analysis of the words of interest concerning two stereotypes (the “Plot 2 stereotypes!” option) and another that enables the analysis of the list of words concerning four stereotypes (the “Plot 4 stereotypes!” option).

When the "Plot 2 stereotypes!" is selected, a bar plot is depicted with the list of words of interest on the y-axis and the stereotypes on the x-axis. The intensity and length of the bar indicate towards which extreme of bias the words of interest are associated. %\todo[inline]{Aca agregaria una referencia a una figura de otra seccion, pero no sabia si estaba ok}
%\todo[inline]{revisar el siguiente párrafo}
When the "Plot 4 stereotypes!" is selected (Figure~\ref{fig:visual-words-bias}B), each word of interested is plotted on the plane. The position of the word indicates towards which extreme of bias the word is more related to. For example, in Figure~\ref{fig:visual-words-bias}~B the word "pelear" (that means to fight in spanish) is more biased towards with man and old stereotypes, whereas the word "maquillar" (that means to put on makeup in spanish) is biased towards to the woman and young stereotype.

\subsubsection{Biases in Sentences tab}
\begin{figure*}[ht] 
    \begin{center}
    \includegraphics[width=\linewidth]{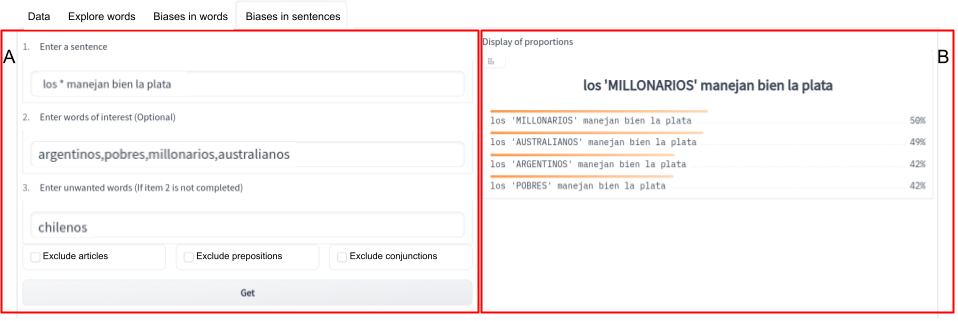}
    \caption{The EDIA visual interface, with the "Biases in Sentences" tab visible. (A) Shows the input panel. Here, the sentence to be evaluated is entered. The sentence must contain a blank, represented by a "*". The list of words of interest and the unwanted words list can be entered in this panel. (B) Displays a list of sentences and their corresponding probabilities, which indicate the likelihood of the sentences being completed with each word of interest based on the language model. If no words of interest are entered, the tool will fill the blank in the sentence with the first five words that the language model considers more likely to occupy the space. In the case that unwanted words were entered, these are not considered in the analysis.
    \label{fig:visual-interface-bias-sentences}}
\end{center}
\end{figure*}
The Biases in Sentences tab aims to explore biases in LLMs, using BETO \cite{CaneteCFP2020} as the default language model. 
%This tab enables the analysis of concepts that include more than word (e.g., Latin American)

The main idea in the Biases in Sentences tab is to select a sentence that contains a blank (in the prototype, the blank is represented with a *) and compare the probabilities obtained by the language model by filling the blank with different words. The user is able to select a list of words of interest and a list of unwanted words (Figure~\ref{fig:visual-interface-bias-sentences}A). The words entered in the unwanted word list will be dismissed from the analysis. Also, articles, prepositions and conjunctions can be excluded from the analysis. If the words of interest are entered by the user, the output panel (Figure~\ref{fig:visual-interface-bias-sentences}B) shows a list of sentences with their corresponding probabilities on the right. These probabilities indicate the possibility of the occurrence of the sentences completed with each word of interest according to the language model. When no list of words is entered, the tool will fill the blanks in the initial sentence with the first five words that the language model considers more likely to occupy the space.
%\todo[inline]{Me parecio que no era necesario agregar como se calculan las probabilidades pero puedo agregarlo si lo consideran necesario}

%% file: 5_user_story.tex
\section{User story in an NLP application}
\label{sec:use_case}

%\todo[inline]{convertir el use case en unas recomendaciones o trayecto metodológico, como las etapas o aspectos relevantes en el proceso de caracterización del sesgo en PLN}

Up to this point we have motivated the need for bias assessment in language technologies and in word embeddings and LLMs in particular; we have explained our differences in the framing of the solution with respect to existing tools; we have discussed the unnecessary technical barriers of existing approaches, that hinder the engagement of actual experts in the exploration process; and we presented the software tool that we developed to allow social scientists and data scientists to collaboratively explore biases and define lists of words and lists of sentences that are useful to detect biases in a particular domain. 
In this Section we describe a user story that presents a paradigmatic process of bias exploration and assessment.%, and we finish %in the following Section with concise recommendations for such processes.

%The methodology presented here is aimed to audit NLP systems. A secondary scenario for the tools used to implement this methodology can well be the exploration of corpora. Indeed, neural approaches constitute very potent statistical machinery to inspect corpora. However, the requirements for such exploration diverge from those of an audit, mostly in the efforts required for the collection and curation of the linguistic data. Therefore, we are not taking into account this scenario in these recommendations.\todo{this paragraph maybe move somewhere else, to the beginning of the paper}

We would like to note that this user story was originally developed to be situated in Argentina, the local context of this project. It was distilled from experiences with data scientists and experts in discrimination that are described in Annex \ref{sec:case-studies}. However, in order to make understanding easier for non-Spanish speaking readers, we adapted the case to work with English, and consequently localized the use case as if it had happened in the United States.

\paragraph{The users.} Marilina is a data scientist working on a project to develop an application that helps the public administration to classify citizens' requests and route them to the most adequate department in the public administration office she works for. 
%, or even provide an automatic response to them, if possible. 
Tomás is a social worker within the non-discrimination office, and wants to assess the possible discriminatory behaviours of such software.

\paragraph{The context.} Marilina addresses the project as a supervised text classification problem. To classify new texts from citizens, they are compared to documents that were manually classified in the past. New texts are assigned the same label as the document that is most similar. Calculating similarity is a key point in this process, and can be done in many ways: programming rules explicitly, via machine learning with manual feature engineering or by deep learning, where a key component is word embeddings. Marilina observes that the latter approach has the least classification errors on the past data she separated for evaluation (the so called test set). Moreover, deep learning seems to be the preferred solution these days, it is often presented as a breakthrough for many natural language processing tasks. So Marilina decides to pursue that option.
%\todo{integrate language models}

An important component of the deep learning approach she uses are word embeddings. Marilina decides to try a well-known word embedding, pre-trained on Wikipedia content. When she integrates it in the pipeline, there is a  boost in the performance of the system: more texts are classified correctly in her test set.

\paragraph{Looking for bias.} Marilina decides to look at the classification results beyond the figures of classification precision. Being a descendant of Latin American immigrants, she looks at documents related to this societal group. She finds that applications for small business grants presented by Latin American immigrants or citizens of Latin American descent are sometimes erroneously classified as immigration issues and routed to the wrong department. These errors imply a longer process to address these requests in average, and sometimes misclassified requests get lost. In some cases, this mishap makes the applicant drop the process.

\paragraph{Finding systematic errors.} Intrigued by this behaviour of the automatic pipeline, she makes a more thorough research into how requests by immigrants are classified, in comparison with requests by non-immigrants. As she did for Latin American requests, she finds that documents presented by other immigrants have a higher error rate than the non immigrants requests. She suspects that other societal groups may suffer from higher error rates, but she focuses on Latin American immigrants because she has a better understanding of the idiosyncrasy of that group, and it can help her establish a basis for further inquiry. She finds some patterns in the misclassifications. In particular, she finds that some particular business, like hairdressers or bakeries, accumulate more errors than others.

\begin{figure*}[ht] 
    \begin{center}
    \includegraphics[width=\linewidth]{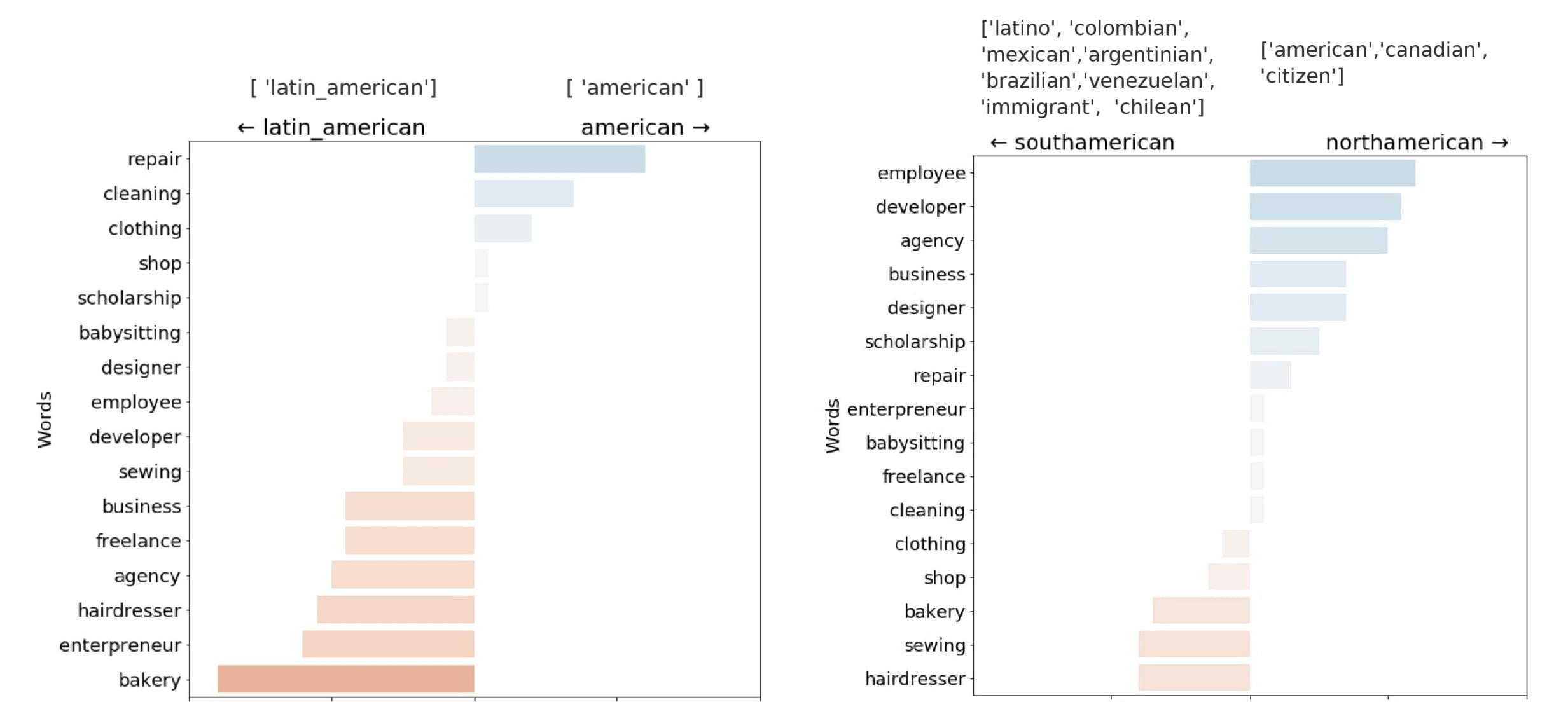}
    \caption{Different characterizations of the space of bias "\textit{Latin American}" vs "\textit{North American}", with different word lists created by a data scientist (left) and a social scientist (right), and the different effect to define the bias space as reflected in the position of the words of interest (column in the left).
    \label{fig:use-case-fig}}
\end{center}
\end{figure*}

\paragraph{Finding the component responsible for bias.} She traces the detail of how such documents are processed by the pipeline and finds that they are considered most similar to other documents that are not related to professional activities, but to immigration. The word embedding is the pipeline component that determines similarities, so she looks into the embedding with the \href{https://github.com/fvialibre/edia}{EDIA toolkit}\footnote{\url{https://github.com/fvialibre/edia}}. She defines a bias space with "\textit{Latin American}" in one extreme and "\textit{North American}" in the other, and checks the relative position of some professions with respect to those two extremes, as can be seen in Figure~\ref{fig:use-case-fig}, on the left. This graph is generated using the button called "Find 2 stereotypes" in the tab. She finds that, as she suspected, some of the words related to the professional field are more strongly related to words related to Latin American than to words related to North American, that is, words like "\textit{hairdresser}" and "\textit{bakery}" are closer to Latin American. However, the words more strongly associated to North American do not correspond to her intuitions. She is at a loss as to how to proceed with this inspection beyond the anecdotal findings, and how to take action with respect to the findings. That is when she calls for help to the non-discrimination office. 

\paragraph{Assessing harm.}%\todo{relate this with the public policy section}
The non-discrimination office appoints Tomás for the task of assessing the discriminatory behavior of the software. Briefed by Marilina about her findings, he finds that misclassifications do involve some harm to the affected people that is typified among the discriminatory practices that the office tries to prevent. Misclassification implies that the process takes longer than for other people, because they need to be reclassified manually before they can actually be taken care of. Sometimes, they are simply dismissed by the wrong civil servant, resulting in unequal denial of benefits. In many cases, the mistake itself has a negative effect on the self-perception of the issuer, making them feel less deserving and discouraging the pursuit of the grant or even the business initiative. Tomás can look at the output of the system, but he cannot see a rationale for the system's misclassifications,  he doesn't know how the automatic classification works.

\paragraph{Detecting the technical barrier.} Tomás understands that there is an underlying component of the software that is impacting in the behaviour of classification. Marilina explains to him that it is a pre-trained word embedding, and that a word embedding is a projection of words from a sparse space where each context of co-occurrence is one of thousands of dimensions into a dense space where there are less dimensions, obtained with a neural network. She says that each word is a vector with numbers in each of those dimensions. Tomás feels that understanding the embedding is beyond his capabilities. Then Marilina explains to him that words are represented as a summary of their contexts of occurrence in a corpus of texts, but this cannot be directly seen, but explored using similarity between words, so that more similar words are closer. 

\paragraph{Finding an intuitive tool for bias exploration.} She shows him some of the tools available to assess bias in the \href{https://huggingface.co/spaces/vialibre/edia}{EDIA demo}\footnote{\url{https://huggingface.co/spaces/vialibre/edia}}, which do not require Tomás to handle any programming or seeing any code. Marilina resorts to the available introductary materials for our tool to explain bias definition and exploration easily to Tomás using the "Biases in words" tab. He quickly grasps the concepts of bias space, definition of the space by lists of words, assessment by observing how words are positioned within that space, and exploration by modifying lists of words, both defining the space and positioned in the space using the "explore words" tab with words that he know are representative for their domain. He gets more insights on the possibilities of the techniques and on possible misunderstandings by reading examples and watching the short tutorials that can be found with the tool. He then understands that word ambiguity may obscure the phenomena that one wants to study if exploring single words, that word frequency has a big impact, and that language-specific phenomena, like grammatical gender or levels of formality, need to be carefully taken into account. He uses the tab "biases in sentences" when words are highly ambiguous or when he needs to express a concept using multiword expressions such as in "Latin America". After some toying with the demo, Tomás believes this tool allows him to adequately explore biases, so Marilina deploys a local instance of the tool, which will allow Tomás to assess the embedding that she is actually using in her development, and the corpus it has been trained on.

\paragraph{Explore the corpus behind the embeddings.} To begin with, Tomás wants to explore the words that are deemed similar to "\textit{Latin American}", because he wants to see which words may be strongly associated to the concept, besides what Marilina already observed. He uses the "data tab" of EDIA, described in Section~\ref{sec:methodology} to explore the data over which the embedding used by Marilina has been inferred. He finds that the embedding has been trained with texts from newspapers. Most of the news containing the word Latin American deal with catastrophes, troubles and other negative news from Latin American countries, or else portray stereotyped Latin Americans, referring to the typical customs of their countries of origin rather than to their facets as citizens in the United States. With respect to business and professions, Latin Americans tend to be depicted in accordance with the prevailing stereotypes and historic occupations of that societal group in the States, like construction workers, waiters, farm hands, etc. He concludes that this corpus, and, as a consequence, the word embedding obtained from it, contains many stereotypes about Latin Americans which are then related to the behaviour of the classification software, associating certain professional activities and demographic groups more strongly with immigration than with business. Marilina says that possibly they will have to find another word embedding, but he wants to characterize the biases first so that he can compare to other word embeddings.

%\paragraph{Cultural aware bias exploration.} Tomás needs to focus his exploration of the word embedding in the bias of interest, in this case, in the Latin American versus North American. 

\paragraph{Formalize a starting point for bias exploration.} Tomás builds lists of relevant words, with the final objective to make a report and take informed action to prevent discriminatory behavior. First, he builds the sets of words that will be representing each of the relevant extremes of the bias space. He realizes that Marilina's approach with only one word in each extreme is not quite robust, because it may be heavily influenced by properties of that single word. That is why he defines each of the extremes of the bias space with longer word lists, and experiments with different lists and how they determine the relative position of his words of interest. Words of interest are the words being positioned in the bias space, words that Tomás wants to characterize with respect to this bias because he suspects that their characterization is one of the causes for the discriminatory behavior of the classification software. 

To find words to include in the word lists for the extremes, Tomás resorts to the functionality of finding the closest words in the embedding. Using "\textit{Latin American}" as a starting point, he finds other similar words like "\textit{latino}", and also nationalities of Latin America using the "Explore Words" tab.

He also explores the contexts of his words of interest. Doing this, he finds that "\textit{shop}" occurs in many more contexts than he had originally imagined, many with different meanings, for example, short for Photoshop. This makes him think that this word is probably not a very good indicator of the kind of behavior in words that he is trying to characterize. He also finds that some professions that were initially interesting for him, like "\textit{capoeira trainer}" are very infrequent and their characterization does not have a correspondence with his intuition about the meaning and use of the word, so he discards them.

Finally, he is satisfied with the definition provided by the word lists that can be seen in Figure~\ref{fig:use-case-fig}, right. With that list of words, the characterization of the words of interest shows tendencies that have a correspondence with the misclassifications of the final system: applications from hairdressers, bakers, dressmakers of latino origin or descent are misclassified more often than applications for other kinds of businesses.

Even though they are assessing biases in a word embedding, that represents words in isolation, collapsing all senses of a word, Tomás believes that once they are characterizing this bias, they may best take advantage of the effort and also build a list of sentences characterizing the same bias, to be used when assessing this same bias in a language model, for example, to assess the behavior of a chatbot. To provide him with inspiration, Marilina offers Tomás a benchmark for bias exploration developed for English and French~\cite{neveol-etal-2022-french} and Tomás uses that dataset partially to define his own list of sentences to explore relevant biases in this domain. 

\paragraph{Report biases and propose a mitigation strategy}. With this characterization of the bias, Tomás can make a detailed report of the discriminatory behavior of the classification system. From the beginning, he suspected the cultural and social reasons behind the errors, which affect more often people of Latin American descent applying for subsidies for a certain kind of business. However, his intuitive manipulation of the underlying word embedding allowed him to find words and phrases that give rise to the pattern of behavior he was observing, going beyond the cases that he has actually been able to see as misclassified by the system, and predicting other cases.

Moreover, understanding the pattern of behavior allowed him to describe properties of the underlying corpus that would be desirable in order to find another word embedding. He can propose strategies like editing the sentences containing hairdressers, designers and bakers to show a more balanced mix of nationalities and ethnicities in them. Finally, he has a list of words and sentences that can give Marilisa to measure and compare the  biases with respect to these aspects in other word embeddings 

%\paragraph{Wider and deeper.}
%If time allows, he will also be able to explore intersectionality and compare this with other embeddings, all in a visual interface with intuitive concepts

%% file: 6_summary.tex
\section{Summary and next steps}

In this paper we have presented a methodology for the kind of involvement that can enrich approaches to bias exploration of NLP artifacts, namely, word embeddings and large language models, with the necessary domain knowledge to adequately model the problems of interest.

We have shown that existing approaches hinder the engagement of experts without technical skills, and focus on metrics and mathematical formalization instead of focusing on the representation of relevant knowledge.

We have presented a methodology to carry out bias assessment in word embeddings and large language models, together with a prototype that aims to facilitate such approach \url{https://huggingface.co/spaces/vialibre/edia}. This methodology addresses bias that can be observed in words in isolation but also in context, which allows to explore both word embeddings and language models, respectively.

The work presented here is just the starting point of a much longer endeavor. Our vision is that firms and institutions integrate this kind of exploration within the development of language technologies, engaging discrimination experts as a permanent asset in their teams, well before deploying any product. We would also like the general population to carry out this kind of audits, and that this is part of a more aware, empowering technology education for all. To achieve that, we envision the following next steps in this line of work:

\begin{itemize}
    \item usability studies and pilots, to find weak spots in the methodology, improve it and adapt it to different scenarios
    \item developing linguistic resources that allow to systematize, and partially automatize, the diagnostic of biases in language artifacts. We are thinking about a repository of lists of words and sentences that capture the linguistic manifestations of different kinds of discrimination, local to different contexts. Developing such resources needs to be a separated effort for different languages and cultures, although a core methodology can be shared.
    \item applying this methodology to multi-language models.
    \item applying the principles in our approach to text-image generation devices, like stable diffusion.
    \item developing a wider set of metrics, that allow to highlight different kinds of diagnostics, including the severity of the discriminative behavior of models.
    \item advocating the adoption of these methodologies within firms and institutions.
    \item sensibilization and education efforts to raise awareness on the problems brought by these technologies, empowering users as capable actors in detecting discriminatory behaviors in language technologies, and getting them to know about systematic methodologies to assess bias.
    \end{itemize}

%% file: case-studies.tex
\section{Two case studies} \label{sec:case-studies}

The main goal of our project is to facilitate access to the tools for bias exploration in word embeddings for people without technical skills. To do that, we explored the usability of our adaptation to Spanish of the Responsibly toolkit\footnote{https://docs.responsibly.ai/}. In Section~\ref{sec:existing-tools} we explain the reasons for using Responsibly as the basis for our work. 

In order to assess where the available tools are lacking and barriers for their use, we conducted two usability studies with different profiles of users: junior data scientists most of them coming from a non-technical background but with a 350 hour instruction in machine learning, and social scientists without technical skills but a 2 hour introduction on word embeddings and bias in language technology. Our objective was to teach these two groups of people how to explore biases in word embeddings, while at the same time gathering information on how they understood and used the technique proposed by \citet{10.5555/3157382.3157584} to model bias spaces. We focused on difficulties to understand how to model the bias space, shortcomings to capture the phenomena of interest and the possibilities the tools offered. 

Our design goals are the following. First, reduction of the technical barrier to a bare minimum. Second, a focus on exploration and characterization of bias (instead of focus on a compact, opaque, metric-based diagnostic). Third, an interface that shows word lists in a dynamic, interactive way that elicits, shapes, and expresses contextualized domain knowledge (instead of taking lists as given by other papers, even if these are papers from social scientists). Fourth, guidance about linguistic and cultural aspects that may bias word lists (instead of just translating word lists from another language or taking professions as neutral).

The first group we studied were students at the end of a 1 year nano-degree on Data Science, totalling 180 hours and a project. All of them had some degree of technical skills, the nano-degree providing extensive practice with machine learning, but most came from a non-computer science background and had not training on natural language processing as part of the course. The second group were journalists, linguists and social scientists without technical skills. 

Both of the groups were given a 2 hour tutorial, based on a Jupyter Notebook we created for Spanish\footnote{Our notebook is available here https://tinyurl.com/ycxz8d9e} as explained in Section~\ref{sec:word-embeddings} by adapting the implementation of the Responsibly toolkit~\cite{10.5555/3157382.3157584} done by Shlomi\footnote{The original notebook for English is available here https://learn.responsibly.ai/word-embedding/}. The groups conducted the analysis on Spanish FastText vectors, trained on the Spanish Billion Word Corpus \cite{cardellinoSBWCE} using a 100 thousand word vocabulary. 

%Both of the groups were given a 2 hour tutorial, based on a Jupyter Notebook created by Shlomi Hod, on how to diagnose bias in word embeddings using lists of words. 
%The technique requires the definition of 2 word lists: 1 for each bias space extreme and a list of words to diagnose. 
%\todo[inline]{explain here the basic methodology, referring to Section sec:word-embeddings}

%Both groups conducted the analysis on Spanish FastText vectors, trained on the Spanish Billion Word Corpus \cite{cardellinoSBWCE} using a reduced version with a 100 thousand word vocabulary\todo{explain here the adaptation to Spanish}.%In Annex X we present a description of each group alongside the lists of words they used to define each bias extreme.

\subsection{Case Study 1: Junior data scientists}

%riqueza-pobreza respecto a lo que quede mejor. sustantivos abstractos o profesiones

The group composed by junior data scientists were given a 1-hour explanation on how the tools were designed and an example of how it could be applied to explore and mitigate gender bias, the prototypical example of application. This was  part of an 8 hour course on Practical Approaches to Ethics in Data Science. As part of the explanation, mitigation strategies built upon the same methodology were also provided, together with the assessment that performance metrics did not decrease in a couple of downstream applications with mitigated embeddings. The presentation of the tools was made for English and Spanish explaining the analogies and differences between both languages to the students, who were mostly bilingual. 
Then, as a take-home activity, they had to work in teams to explore a 2 dimensional bias space of their choice, different from gender.

%\todo{explain how we actually tried to simplify technicalities, as a starting point in our methodology}

%\todo{explain that we provided a *methodology*, very succintly}

During the presentation of the tools, students did not request clarifications or extensive explanations into the nuances of word embeddings, biases represented as lists of words, or the linguistic differences between English and Spanish and the adaptation of the tool. We suspect this was due to the fact that they were conditioned by the methodology of the nano-degree, which was based on classes explaining a methodology and showing how it was applied, followed by practical sessions when they actually applied the methodology to other cases. Thus, they were not trying to be critical, but to reproduce the methodology.

\begin{figure}[ht] 
\begin{center}
\includegraphics[width=\linewidth]{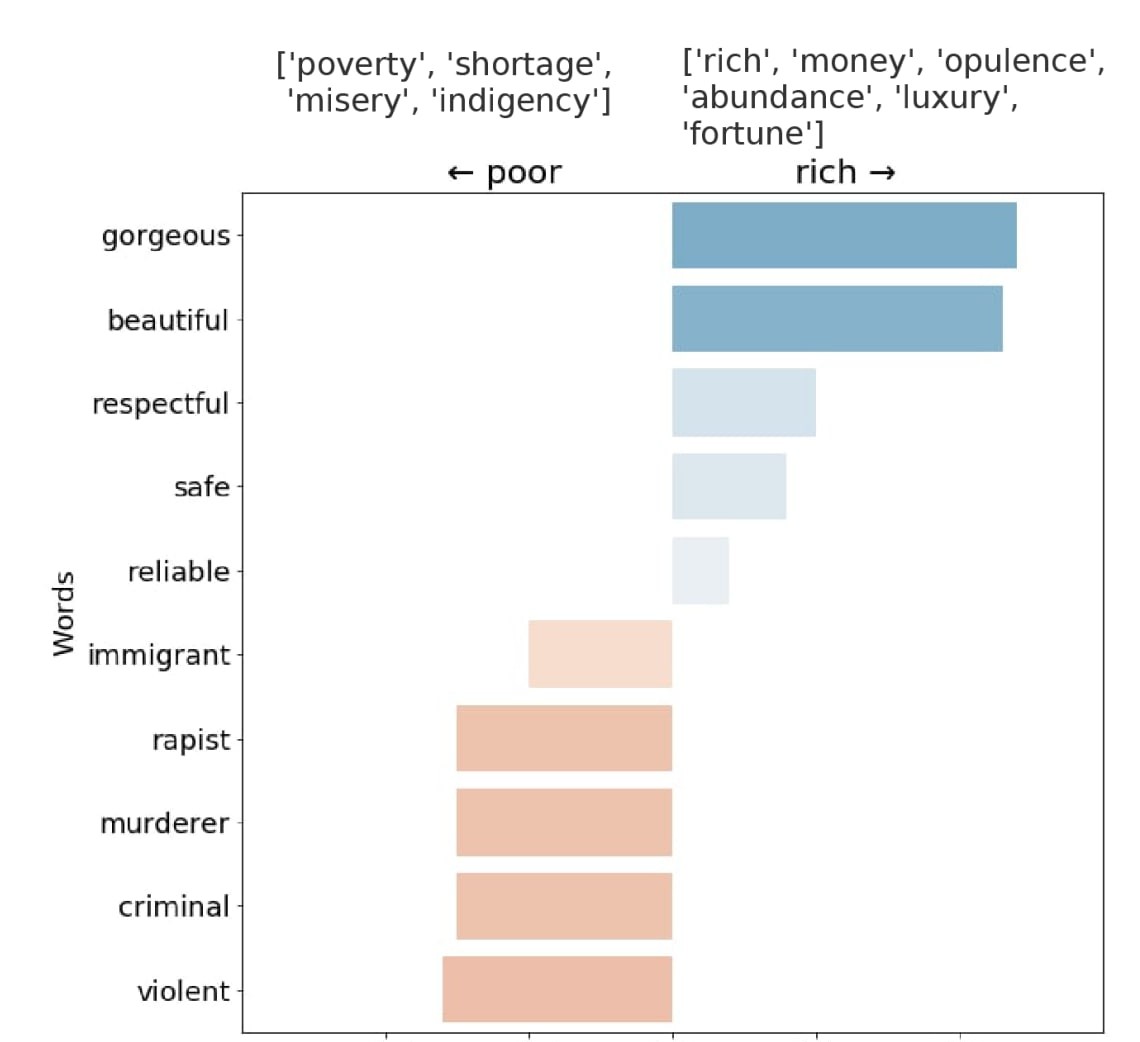}
\caption{Exploration of the \textit{rich} vs \textit{poor} bias space carried out by data scientists, showing the words used to define the two extremes of the bias space and how words of interest, like "\textit{gorgeous}" or "\textit{violence}", are positioned with respect to each of the extremes.  The original exploration was carried out in Spanish and has been translated into English for readability. \label{fig:data-scientists}}
\end{center}
\end{figure}

The teams successfully applied the methodology to characterize biases other than gender, namely economic bias (wealthy vs poor), geographical bias (latin american vs north american), ageism (old vs young). Figure~\ref{fig:data-scientists} illustrates one of their analyses of the bias space defined by \textit{rich} vs \textit{poor}, exploring how negative and positive words were positioned in that space. The figure shows the list of words they used to define the two extremes of the bias space, the concepts of poor and rich. %developed by the students in the data science course. 
It also shows how words %a graphical representation of which words, selected by the students, 
such as \textit{gorgeous} and \textit{violence} are closer to the rich or poor concepts.% as defined by them. 
Students concluded that the concept of poor is more associated with words with a negative sentiment and rich more with a positive sentiment.

They did not report major difficulties or frustrations, and in general reported that they were satisfied with their findings by applying the tool. They also applied mitigation strategies available with the Responsibly toolkit, but they made no analysis of its impact. 

As it was not required by the assessment, they did not systematically report their exploration process. To gain further insights on the exploration process, we included an observation of the process in the experience with social scientists.

Overall, their application of the tool was satisfactory but rather uncritical. This is not to suggest that the participants were uncritical themselves, but rather that the way the methodology was presented, aligned with a consistent approach to applying methodologies learnt throughout the course, inhibited a more critical, nuanced exploitation of the tool.

\subsection{Case Study 2: Social Scientists}
The social scientists were presented the tool as part of a twelve-hour workshop on tools to inspect NLP technologies. A critical view was fostered, and we explicitly asked participants for feedback to improve the tool.

There was extensive time within the workshop devoted to carry out the exploration of bias. We observed and sometimes elicited their processes, from the guided selection of biases to be explored, based on personal background and experience, to the actual tinkering with the available tools. Also, explicit connections were made between the word embedding exploration tools available via Responsibly and an interactive platform for NLP model understanding, the Language Interpretability Tool (LIT) \cite{Tenney2020}, which was also inspiring for participants as to what other information they could obtain that could enrich their analysis in exploration.

Since there was no requirement for a formal report, bias exploration was not described systematically. Different biases were explored, in different depths and lengths. Besides gender and age, also granularities of origin (cuban - north-american) and the intersection between age and technology were explored. Participants were creative and worked collaboratively to find satisfactory words that represented the phenomena that they were trying to assess. They were also insightful in their analysis of the results: they were able to discuss different hypotheses as to why a given word might be further in one of the extremes of the space than another. 

\begin{figure}[ht] 
\begin{center}
\includegraphics[width=\linewidth]{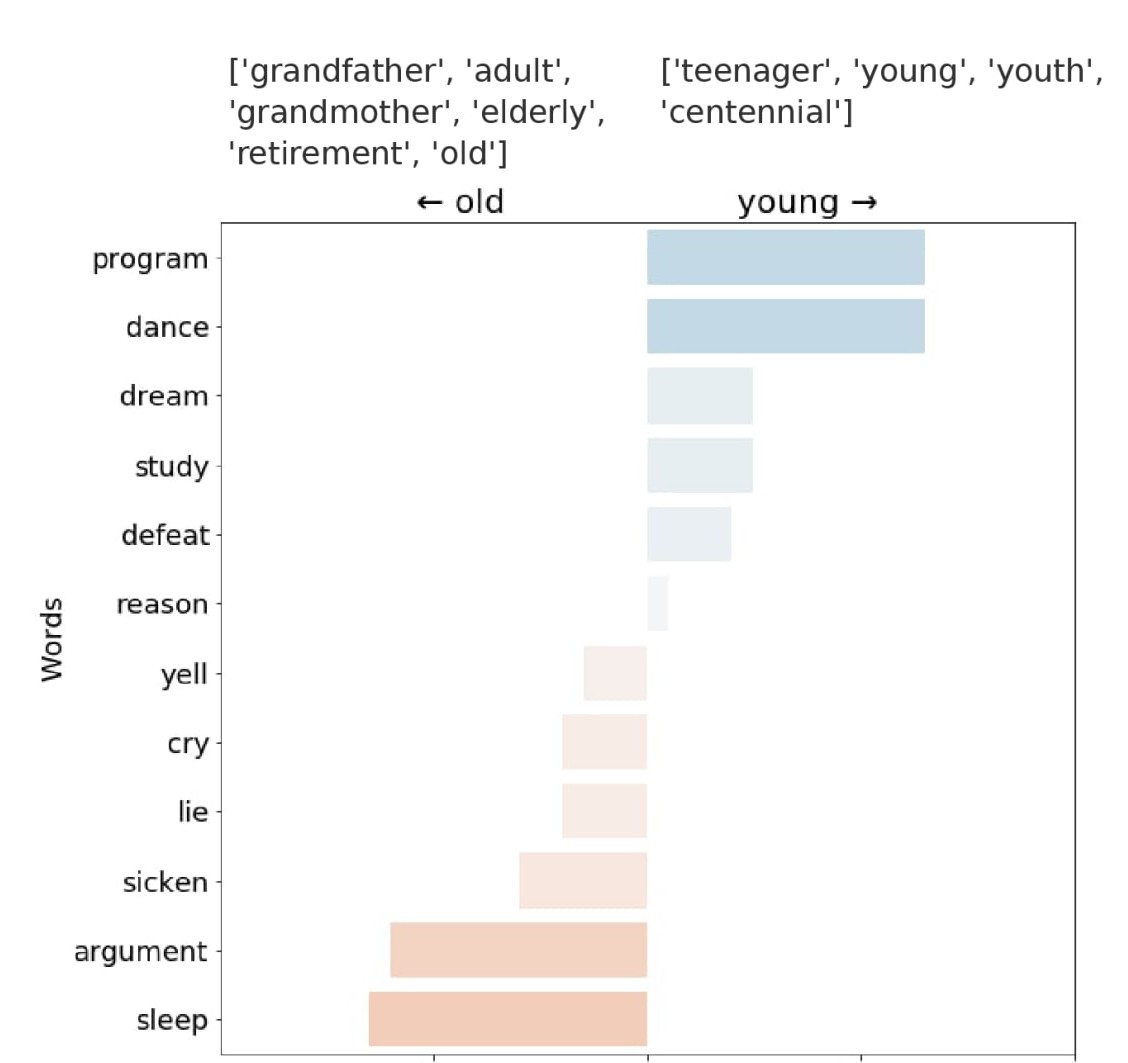}
\caption{Exploration of the \textit{old} vs \textit{young} bias space carried out by social scientists, showing the words used to define the two extremes of the bias space and how words of interest, like "\textit{dance}" or "\textit{sleep}", are positioned with respect to each of the extremes. The original exploration was carried out in Spanish and has been translated into English for readability. \label{fig:social-scientists}}
\end{center}
\end{figure}

Figure~\ref{fig:social-scientists} illustrates one of their analysis of the bias space defined by the concepts of \textit{young} and \textit{old}, using the position of verbs in this space to explore bias. In this analysis they did not arrive to any definite conclusions, but found that they required more insights on the textual data wherefrom the embedding had been inferred. For example, they wanted to see actual contexts of occurrence of "\textit{sleep}" or "\textit{argument}" with "\textit{old}" and related words, to account for the fact that they are closer to the extreme of bias representing the "\textit{old}" concept. Analyzing these results, the group also realized there were various senses associated to the words representing the "old" concept, some of them positive and some negative. They also realized that the concept itself may convey different biases, for example, respect in some cultures or disregard in others. Such findings were beyond the scope of simple analysis of the embedding, requiring more contextual data to be properly analyzed and subsequently addressed.

Participants were active and creative while requiring complementary information about the texts from which word embeddings had been inferred. Many of these requirements will be included in the prototype of the tool we are devising, such as the following:
frequencies of the words being explored
concordances of the words being explored, that is, being able to examine the actual textual productions
when building word lists, especially for the extremes that define the bias space, suggestions of similar words or words that are close in the embedding space would be useful, as it is often difficult to come up with those and the space is better represented if lengthier lists are used to describe the extremes
functionalities and a user interface that facilitates the comparison between different delimitations (word listst) of the bias space, or the same delimitations in different embeddings. 

They also stated that it would be valuable that the tool allowed them to explore different embeddings, from different time spans, geographical origin, publications, genres or domains. For that purpose, our prototype includes the possibility to upload a corpus and have embeddings inferred from that corpus, which can then be explored.

In general, we noted that social scientists asked for more context to draw conclusions on the exploration. They were critical of the whole analysis process, with declarations like ``I feel like I am torturing the data''. Social scientists study data in context, not data in itself, as is more common in the practice of data scientists. Also, the context where the tool was presented also fostered a more critical approach. We asked participants to formulate the directions of their explorations in terms of hypotheses. Such requirement made it clear that more information about the training data was needed in order to formulate hypotheses more clearly.

%\subsection{Wishlist for the prototype}

%In these two case studies we could test the adequacy of some aspects of our approach, describe the shortcomings of the tools that we are currently using at this stage of development, and also reassess some design decisions with respect to the proposed tool.

%Summing up, below we highlight those limitations that we have identified during this case studies that we will address in the prototype. We then discuss intrinsic limitations of the current capabilities of the tools for bias exploration. 

%\noindent
%\framebox{
%\begin{minipage}{0.93\columnwidth}
%Working with \textbf{multiword expressions} as a linguistic unit.
%\end{minipage}
%}

% \noindent
% \framebox{
% \begin{minipage}{0.93\columnwidth}
% Any word can be mapped into the embedding space.
% \end{minipage}
% }

\noindent
\framebox{
\begin{minipage}{0.93\columnwidth}
Being able to retrieve the actual \textbf{contexts of occurrence} of words in the corpus wherefrom the embedding has been obtained.
\end{minipage}
}

\subsection{Conclusions of the case studies}

With these study cases, we show that reducing the technical complexity of the tool and explanations to the minimum fosters engagement by domain experts. Providing intuitive tools like word lists, instead of barriers like vectors, allows them to formalize their understanding of the problem, casting relevant concepts and patterns into those tools, formulating hypotheses in those terms and interpreting the data. Such engagement is useful in different moments in the software lifecycle: error analysis, framing of the problem, curation of the dataset and the artifacts obtained.

%\todo[inline]{insert the following in the text}

%\noindent
%\framebox{
%\begin{minipage}{0.93\columnwidth}
%\textbf{Lesson learned}: 
%Our conclusion is that the inspection of biases in word embeddings can be understood without most of the underlying technical detail. However, the Responsibly toolkit addapted to Spanish needs the improvements we discuss in this section and develop as an applied research plan in Section~\ref{sec:next-steps}.
Our conclusion is that the inspection of biases in word embeddings can be understood without most of the underlying technical detail. However, the Responsibly toolkit adapted to Spanish needs the improvements we discuss in this section.